\documentclass[10pt,twocolumn,letterpaper]{article}
\usepackage[pagenumbers]{cvpr} % To force page numbers, e.g. for an arXiv version

\newcommand\blfootnote[1]{%
  \begingroup
  \renewcommand\thefootnote{}\footnote{#1}%
  \addtocounter{footnote}{-1}%
  \endgroup
}

\usepackage{graphicx}
\usepackage{float}
\usepackage{caption}
\usepackage{lipsum}
\usepackage[final]{microtype}
\usepackage{bm}
\usepackage[dvipsnames]{xcolor}

\usepackage[ruled,noend]{algorithm2e} % For algorithms

\SetAlFnt{\small}
\SetAlCapFnt{\small}
\SetAlCapNameFnt{\small}
\SetAlCapHSkip{0pt}
\usepackage{graphics}
\SetKwInput{KwInput}{Input}                % Set the Input
\SetKwInput{KwOutput}{Output}              % set the Output
\SetArgSty{textnormal}

\newcommand{\moniker}{Gaussian Garments}

\definecolor{cvprblue}{rgb}{0.21,0.49,0.74}
\usepackage[pagebackref,breaklinks,colorlinks,citecolor=cvprblue]{hyperref}

\usepackage{paralist}
\usepackage{amsmath}

\usepackage{soul}

\title{\moniker: Reconstructing Simulation-Ready Clothing with Photorealistic Appearance from Multi-View Video}

\author{Boxiang Rong$^{*1}$~~
\and
Artur Grigorev$^{*1,2}$~~
\and
Wenbo Wang$^{1}$~~
\and
Michael J. Black$^{2}$~~
\and
Bernhard Thomaszewski$^{1}$~~
\and
Christina Tsalicoglou$^{1}$~
\and
Otmar Hilliges$^{1}$~
\vspace{0.1em}
\and
~~~~~~~~$^{1}$Department of Computer Science, ETH Zurich~~~~~~~~
\and
~~~~~~~~$^{2}$Max Planck Institute for Intelligent Systems, Tübingen~~~~~~~~
\\
{\small\url{https://ribosome-rbx.github.io/Gaussian-Garments}}
}

\begin{document}
\twocolumn[{
\maketitle
\begin{center}
    \captionsetup{type=figure}
    \makebox[\textwidth][c]{
    \includegraphics[width=1.1\textwidth]{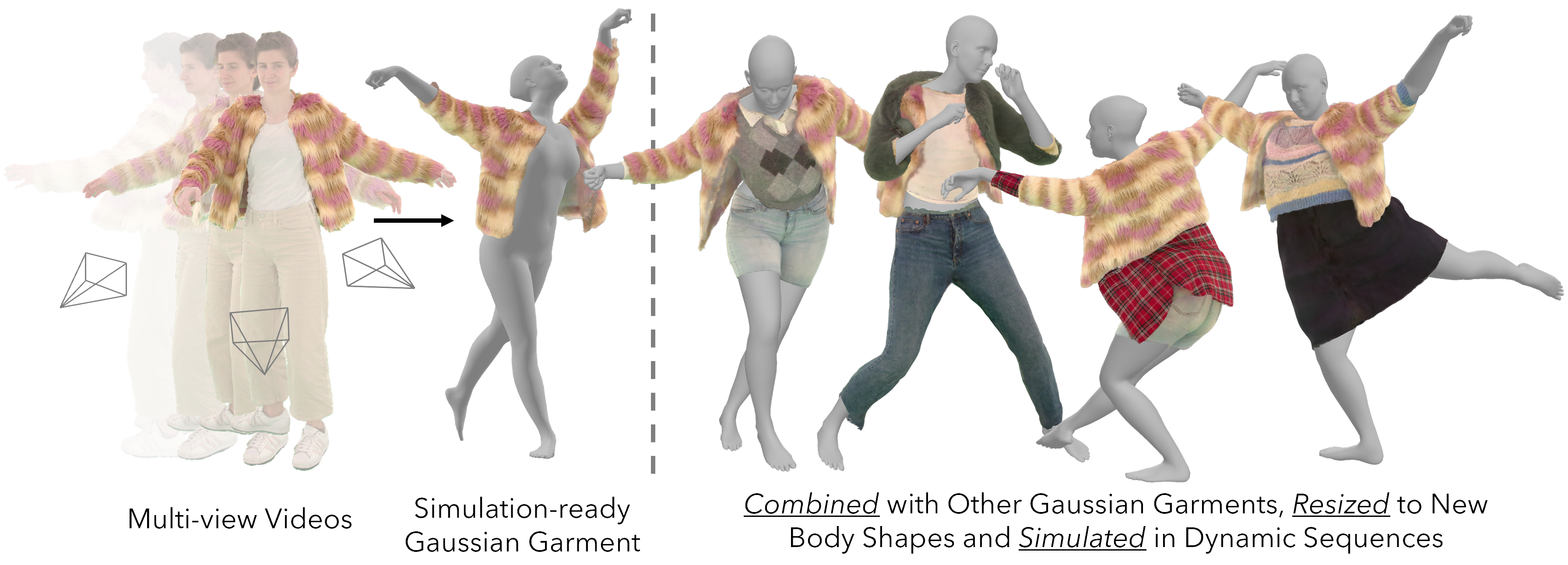}
    }
    \captionof{figure}{
        We introduce \moniker, a novel approach for reconstructing realistic simulation-ready garments from multi-view videos. 
        Our natural coupling of 3D meshes and 3D Gaussian splatting allows {\em Gaussian Garments} to accurately represent both the overall geometry and the high-frequency details of human clothing.
        The reconstructed garments can then be retargeted to novel human models, resized to fit novel body shapes, and simulated over moving bodies with novel motions.
        Our approach also enables the automatic construction of complex multi-layer outfits from a set of separately captured Gaussian garments.      
    }
    \label{fig:teaser}
\end{center}
}]

\blfootnote{$^*$Authors contributed equally}

\begin{abstract}

We introduce \moniker, a novel approach for reconstructing realistic simulation-ready garment assets from multi-view videos. 
Our method represents garments with a combination of a 3D mesh and a Gaussian texture that encodes both the color and high-frequency surface details.
This representation enables accurate registration of garment geometries to multi-view videos and helps disentangle albedo textures from lighting effects.
Furthermore, we demonstrate how a pre-trained graph neural network (GNN) can be fine-tuned to replicate the real behavior of each garment.
The reconstructed Gaussian Garments can be automatically combined into multi-garment outfits and animated with the fine-tuned GNN.
\end{abstract}
\section{Introduction}
Reconstructing and animating human apparel is essential for many applications, from virtual try-on systems to movies and video games.

Faithful digital representation of real garments requires capturing three key aspects. 
First, the 3D \textit{geometry} of the garments must be reconstructed to model both their overall structure and fine details.
Second, the \textit{appearance} of the garments must be recreated to accurately reflect their color and texture.
Finally, the real \textit{behavior} of the garments must be mimicked to produce convincing animations.
Our method, \moniker, leverages the expressivity of 3D Gaussian splatting to reconstruct these three critical aspects from multi-view videos.

In computer graphics, garments are traditionally represented as polygonal meshes with 2D textures. 
While this representation enables efficient simulation and appealing rendering, creating detailed garment meshes is labor-intensive, particularly for complex textures like fur. 
Further, meshes are not well-suited for differentiable optimization of their structure and topology from images. 
To overcome these limitations, recent works have started to explore neural implicit representations (NIRs) as a basis for modeling photorealistic clothing. While NIRs provide strong flexibility in terms of clothing topology and appearance, using them to generate physically realistic motions is exceedingly difficult.

3D Gaussian splatting has recently emerged as a highly efficient and flexible alternative for photorealistic scene reconstruction. Unlike NIRs, Gaussians can be edited individually to accommodate changes in scene geometry, appearance, and lighting. Recent works leverage this ability to generate photorealistic digital copies of clothed humans. However, these methods construct holistic avatars without the ability to extract individual garments as separate assets.
Consequently, they cannot retarget these garments to different bodies, adjust their size, or combine clothing items from various avatars into a novel outfit---tasks crucial for many computer graphics applications.

In this work, we introduce  \moniker---the first method that uses 3D Gaussian splatting to reconstruct photorealistic, simulation-ready assets of human clothing.
At its core, our method combines mesh-based geometry with Gaussian-based appearance modeling.
Starting with an initial garment mesh obtained from multi-view images we register it to a set of multi-view videos using a photometric optimization procedure based on Gaussian splatting. 
Then, we optimize a Gaussian texture to recover the garment's detailed appearance, with disentangled ambient color and view-dependent properties. Finally, using the registered mesh, we fine-tune a graph neural network (GNN) for neural simulation to match the garment's real-world behavior. 

In summary, our main contributions are
\begin{compactitem}
\item a comprehensive pipeline for reconstructing the shape, appearance, and behavior of real-world garments using Gaussian splatting,
\end{compactitem}
\begin{compactitem}
\item an algorithm for registering garment meshes to multi-view videos with an optimization procedure based on Gaussian splatting, and
\end{compactitem}
\begin{compactitem}
\item a Gaussian Garment representation that combines triangle meshes with Gaussian textures to capture photorealistic appearance and can be used as a fully controllable 3D asset.
\end{compactitem}
\section{Related work}

\begin{figure*}
    \centering
    \includegraphics[width=\linewidth]{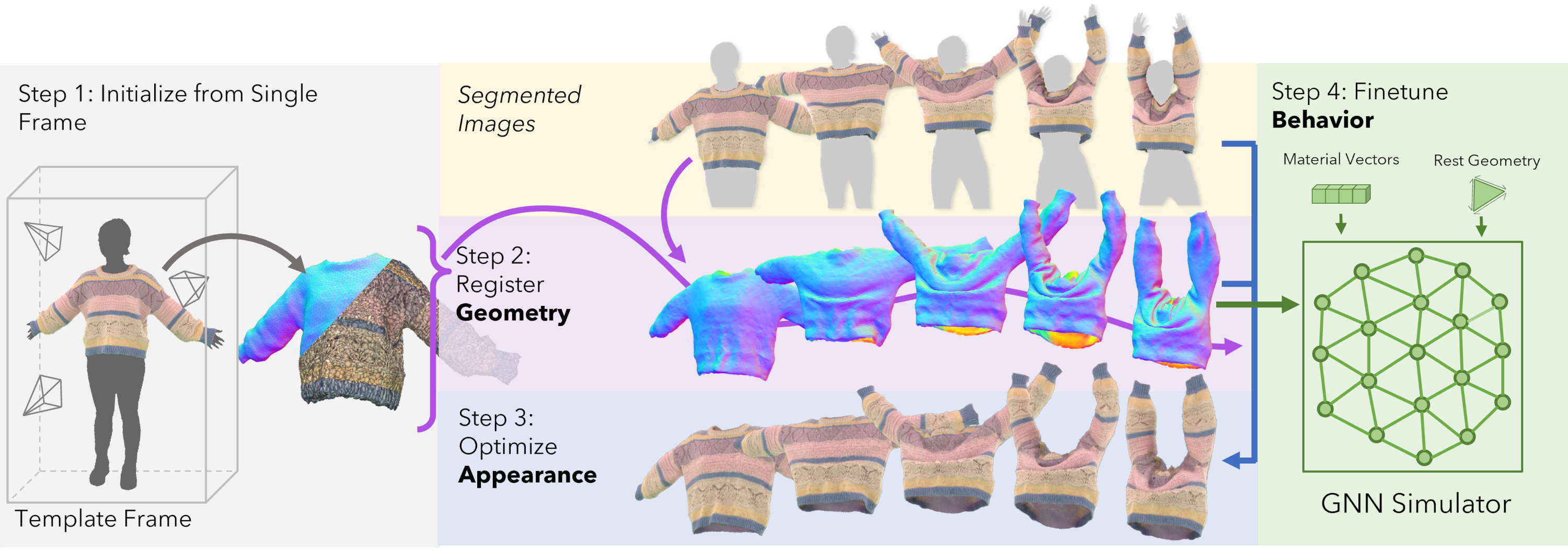}
      \caption{
      The procedure for obtaining simulation-ready photorealistic garment assets consists of four steps. 
      In Step 1, we initialize the garment's geometry and appearance from a single multi-view frame (Sec.~\ref{sec:initialize}). 
      In Step 2, we register the garment geometry to multi-view videos (Sec.~\ref{sec:registration}). 
      In Step 3, we optimize the garment's appearance over the training sequences. 
      In Step 4, we fine-tune a simulation GNN to accurately replicate the garment's real behavior. 
      The resulting garment assets can be directly simulated with the GNN, combined into multi-garment outfits, and resized to fit different body shapes.S
      }
      \label{fig:method}
\end{figure*}
\subsection{Garment reconstruction}
Reconstructing 3D representations of real-world garments is a long-studied task.
Input data in this problem ranges from 4D scans and multi-view videos to single images.

ClothCap \cite{pons2017clothcap} and SIZER~\cite{tiwari2020sizer} segment 4D scans to extract garment meshes, which can be retargeted to novel body shapes and poses. Bang et al.~\cite{bang2021estimating} and NeuralTailor~\cite{NeuralTailor2022} estimate sewing patterns from static 3D scans and point clouds, respectively. These sewing patterns can then be draped over body geometry to produce a 3D mesh.
DiffAvatar~\cite{li2024diffavatar} employs static 3D representation to jointly optimize both the garment's 2D pattern and material properties, resulting in simulation-ready meshes.

BCNet~\cite{jiang2020bcnet} and SMPLicit~\cite{corona2021smplicit} train neural networks to generate template-mesh displacements and unsigned neural fields, respectively, from monocular images. DeepFashion3D~\cite{zhu2020deep} and Zhu et al.~\cite{zhu2022registering} use joint explicit--implicit representations to register garment templates to 2D images. 
However, these methods do not reconstruct the garment's appearance.

SCARF~\cite{feng2022capturing} uses monocular videos to optimize an articulated neural radiance field (NeRF). While it can model garments' appearance over novel body shapes and poses, it suffers from the choice of representation. NeRF reconstructions produce poor geometries and are affected by slow optimization and rendering speed. 
Additionally, SCARF does not allow combining different reconstructed garments.

Closest to our approach are the works by Xiang et al.~\cite{xiang2021modeling, xiang2022dressing}.
They reconstruct textured garment meshes from multi-view videos, with~\cite{xiang2021modeling} also using a physical simulator to generate cloth dynamics. While achieving high visual quality, they use simple textured meshes to represent garments, which limits their ability to model high-frequency geometric details like fur. 
Moreover, they do not provide a means to reconstruct material parameters for the garments and select them manually instead.

With \moniker{}, we demonstrate how 3D garment meshes can be combined with 3D Gaussian splatting technique to achieve photorealistic garment appearance.
Additionally, we fine-tune a garment-modeling GNN to accurately replicate the real garment behavior.

\subsection{3D Gaussian splatting for human avatars}
The recently proposed 3D Gaussian splatting (3DGS) technique~\cite{kerbl3Dgaussians} reconstructs scenes using explicitly defined 3D Gaussian kernels.
This method combines the advantages of both implicit and explicit 3D representations. Similar to Neural Radiance Fields (NeRFs)~\cite{srinivasan2020nerf, barron2021mip} and neural signed distance fields~\cite{wang2021neus, park2019deepsdf}, the 3DGS representation can be optimized from multi-view images and is capable of flexibly modeling diverse topologies.
Additionally, the explicit nature of 3DGS enables it to easily represent dynamic scenes~\cite{luiten2024dynamic} and model physical behavior~\cite{xie2024physgaussian}.

Despite its recent introduction, 3D Gaussian splatting has been adopted by numerous approaches to represent humans in digital environments. 
GaussianAvatar\textbf{s}~\cite{qian2024gaussianavatars} and SplattingAvatar~\cite{shao2024splattingavatar} use parametric meshes with rigidly attached 3D Gaussians to represent human heads and clothed bodies. 
GaussianAvatar~\cite{hu2024gaussianavatar} and 3DGS-Avatar~\cite{qian20243dgsavatar} optimize a canonical Gaussian body, a skinning model, and a neural network that predicts pose-dependent offsets to the Gaussian parameters. 
AnimatableGaussians~\cite{li2024animatable} construct a person-specific canonical template and predict a Gaussian texture containing appearance and geometry parameters. 
The canonical template and diffused skinning model allow~\cite{li2024animatable} to better model loose garments.
PhysAvatar~\cite{zheng2024physavatar} uses 3D Gaussian splatting to register meshes of clothed humans to multi-view videos. It then uses inverse rendering to reconstruct the mesh textures and inverse physics to recover material parameters.
For its final representation, PhysAvatar discards 3D Gaussians and uses flat-textured meshes instead.
This enables relighting the meshes with standard techniques but does not allow the modeling of non-flat surfaces like fur.
Moreover, PhysAvatar does not provide a means to reconstruct template meshes for clothed humans and uses ground-truth ones instead.

A common drawback of these methods is their focus on reconstructing holistic avatars of clothed humans without separating garments from the bodies.
This limitation reduces their applicability in common computer graphics tasks such as simulating garments over different human models, combining garments into outfits, and fitting garment sizes to varying body shapes. 
D3GA~\cite{zielonka2023drivable} addresses this issue by separating garments from human bodies, but it is limited to modeling simple, tight-fitting outfits consisting of two garments (e.g., a T-shirt and pants).

In contrast, Gaussian Garments reconstruct distinct 3D garment assets that can be resized and combined into multi-layer outfits. Fine-tuning a cloth simulation GNN enables realistic modeling of loose garments in dynamic motions.
\section{Method}
We use a set of multi-view videos to reconstruct geometry, appearance, and behavior of a real-world garment. 
Our pipeline, outlined in Fig.~\ref{fig:method}, consists of four main stages. 
First, we initialize Gaussian garment geometry and appearance from a single multi-view frame (Sec.~\ref{sec:initialize}).
Second, we register the garment's geometry to all available frames (Sec.~\ref{sec:registration}). 
Third, we optimize the garment's appearance by disentangling the albedo Gaussian texture from lighting effects and per-frame local transformation offsets predicted by a neural network (Sec.~\ref{sec:appearance}). 
Finally, we fine-tune the garment's behavior by optimizing a graph neural network (GNN)~\cite{grigorev2024contourcraft} to replicate the registered garment motion (Sec.~\ref{sec:behavior}). 
In this section, we detail each of these steps.

Note that apart from RGB frames, our pipeline requires 2D semantic segmentation maps and parametric human body models~\cite{pavlakos2019expressive} fitted to each frame.
We extract these priors from multi-view videos automatically using existing methods.

\subsection{Gaussian garment initialization}
\label{sec:initialize}
\subsubsection{Mesh reconstruction}
As an initial step, we reconstruct the static geometry of a given garment. 
For that, we select a ``template'' multi-view frame where the garment's is surface fully visible.
We recover the garment's 3D mesh from this frame, using existing algorithms for multi-view stereo~\cite{schoenberger2016sfm}, surface reconstruction~\cite{kazhdan2006poisson}, and remeshing~\cite{levy2013variational} (see Sec.~\ref{sec:sm_meshinit}).

Together with the Gaussian texture, described below, the meshes obtained in this step can represent both the overall garment geometry and high-frequency details like fur.

\subsubsection{Gaussian texture}
\label{sec:gtex}
To represent the garment's appearance, we use a so-called Gaussian texture. 
Similar to a traditional texture, it maps between the 3D mesh surface and a 2D texture image that controls the surface appearance.
However, in our case, each point on the texture defines parameters for a 3D Gaussian: spherical harmonic coefficients $\boldsymbol{\phi} \in [0,1]^{16\times3}$, opacity $\alpha$, scale $\mathbf{s} \in \mathbb{R}_{+}^{3}$, local rotation $\mathbf{r} \in \mathbb{H}$ and translational offsets $\boldsymbol{\mu} \in \mathbb{R}^3$.
The latter two are set in a local coordinate frame which we define later. 

\begin{figure}
    \centering
    \includegraphics[width=\linewidth]{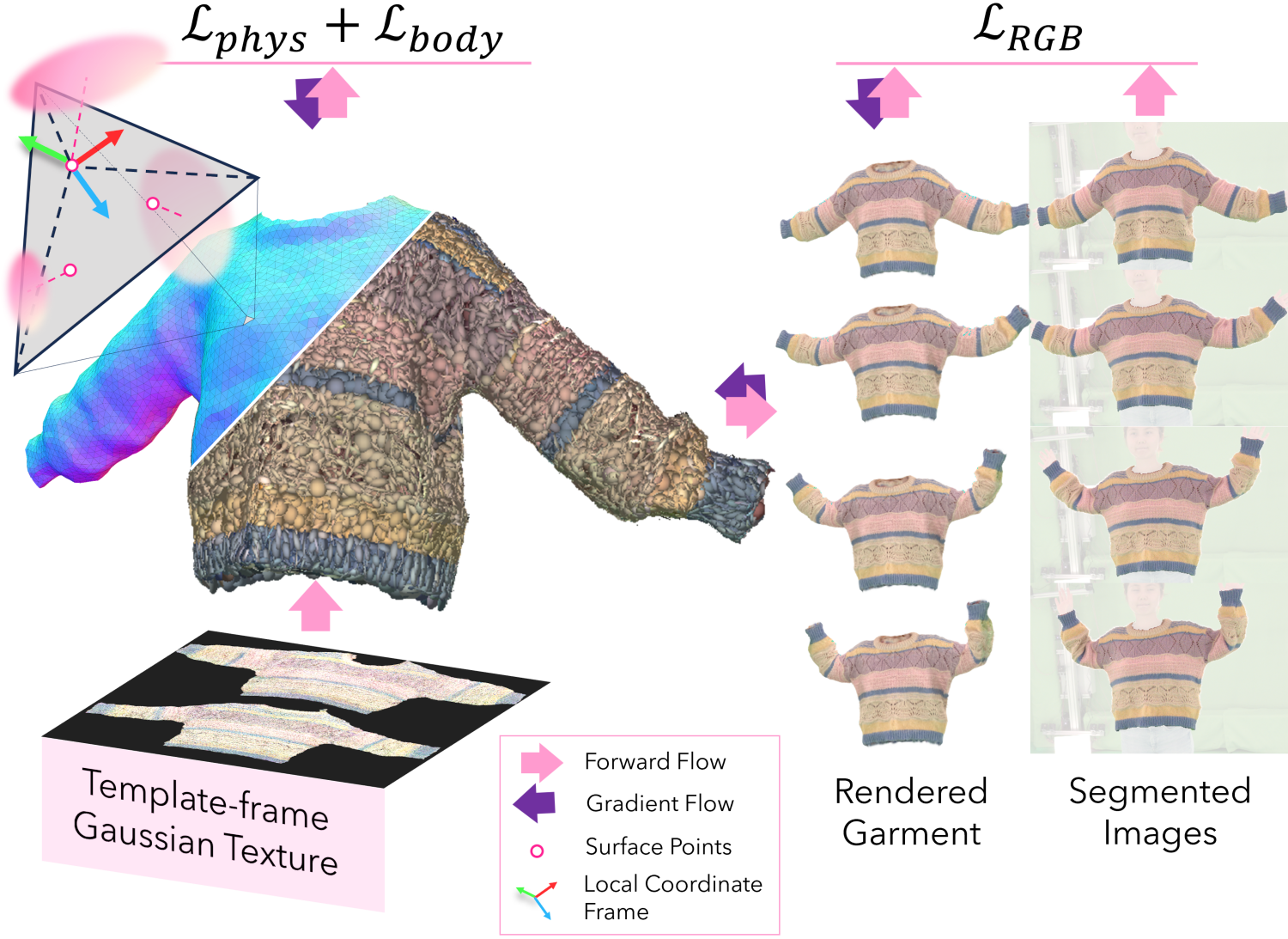}
      \caption{
      To register the garment mesh we render the Gaussians rigidly attached to the mesh faces (top left) and optimize a combination of the RGB loss $\mathcal{L}_{\textit{RGB}}$ and physical energies $\mathcal{L}_{\textit{phys}}$. We also use a body penetration term $\mathcal{L}_{\textit{body}}$ to ensure that the garment conforms to the body model. 
      }
      \label{fig:registration}
\end{figure}

We use the Gaussian texture and the mesh geometry to construct a Gaussian garment in 3D space in the following way.
We first sample the Gaussians from the texture in a regular grid (e.g., once per texel).
The Gaussian's location on the texture controls which 3D face $f_i$ it is attached to and what its barycentric coordinates within $f_i$ are.
These two elements define the initial position of the Gaussian on the mesh surface.
We call this position the Gaussian's ``surface point''.
This surface point serves as the origin for the Gaussian's local coordinate frame.
The basis of this coordinate frame consists of the normal vector for the face $f_i$ and two orthogonal vectors on its surface (see Fig.~\ref{fig:registration}, left).

Following Qian et al.~\cite{qian2024gaussianavatars}, we determine the Gaussian's final 3D position and shape using its scale $\mathbf{s}$, rotation quaternion $\mathbf{r}$, and translational offsets $\boldsymbol{\mu}$. See Sec.~\ref{sec:sm_appearance_init} for details.

\subsection{Tracking-based registration}
\label{sec:registration}

\begin{figure*}
    \centering
    \includegraphics[width=1\linewidth]{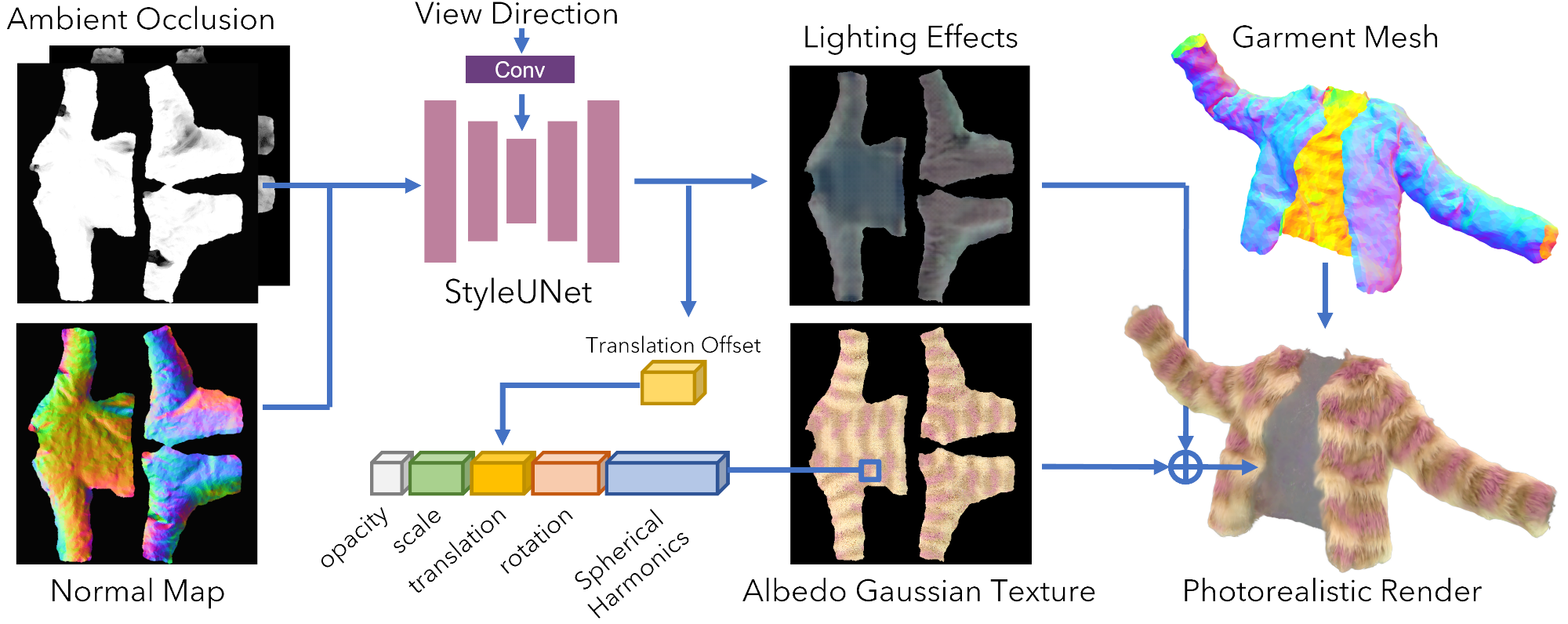}
  \caption{
  We model the appearance of Gaussian Garments using a combination of an albedo Gaussian texture and a neural network that predicts lighting effects and local translational offsets. The albedo Gaussian texture stores color information along with Gaussian parameters, including local rotation, translation, and scale.
  During rendering we regularly sample the Gaussian texture and spawn the 3D Gaussians rigidly attached to the garment surface.      
  }
      \label{fig:appearance}
          
\end{figure*}
To use Gaussian Splatting for geometry registration, we first have to construct an initial appearance model represented by 3D Gaussians. 
To do so, we initialize the Gaussian texture with default parameters, create Gaussians on the mesh surface, and optimize them to match the template-frame observations (see Sec.~\ref{sec:sm_appearance_init} for details). 
This initial appearance model is only used for mesh registration. 
We enhance its visual quality and disentangle albedo color from lighting effects in later steps (Sec.~\ref{sec:appearance}). 
After obtaining a template garment mesh and an initial appearance model, we register the template mesh to multi-view videos.
The key to this process is propagating the gradient from the image space to the positions of the mesh nodes. 
To achieve this, we compute the error $\mathcal{L}_{\textit{RGB}}$ between the rendered Gaussian splats and the ground-truth images. We then pass its gradients through the 3D Gaussians, rigidly attached to the garment's faces, to the nodes of the garment mesh. $\mathcal{L}_{\textit{RGB}}$ is defined as
\begin{align}
\mathcal{L}_{\textit{RGB}} = \lambda_{\textit{RGB}}\mathcal{L}_1 + (1-\lambda_{\textit{RGB}})\mathcal{L}_{\textit{SSIM}},
\end{align}
where $\mathcal{L}_1$ is a mean absolute error, $\mathcal{L}_{\textit{SSIM}}$ is a structural similarity loss, and $\lambda_{\textit{RGB}}$ is a balancing weight.

However, na\"ive minimization of the RGB discrepancy $\mathcal{L}_{\textit{RGB}}$ between renders and observations would result in severely disfigured meshes (see Fig.~\ref{fig:regabl}).
Therefore, we expand the optimized loss function with a set of physical energies.

First, we regularize the angle between each pair of neighboring faces with bending energy 
$\mathcal{L}_{\textit{bending}}$:
\begin{align}
\mathcal{L}_{\textit{bending}} = 
\sum_{(i,j)} \dfrac{\|e_{ij}\|^2}{a_{ij}} 
\mathrm{atan2}(\mathrm{sin}(\theta_{ij}), \mathrm{cos}(\theta_{ij}))^2,
\end{align}
where $(i,j)$ are indices of neighboring triangles, 
$\theta_{ij}$ is the angle between the triangles' normal vectors,
$\|e_{ij}\|$ is the length of the edge connecting the two triangles, and $a_{ij}$ is the sum of their areas.

Second, we regularize the stretching of the triangles relative to the template frame using the strain energy $\mathcal{L}_{\textit{strain}}$, based on the St.~Venant--Kirchhoff material model.
This energy uses the deformation gradient $\mathbf{F} = \dfrac{\partial x_{t}}{\partial X}$ of the current frame geometry $x_{t}$ relative to the template geometry $X$, and is computed as a sum over all faces $f_i$:
\begin{align}
\mathcal{L}_{\textit{strain}} = \sum_{i} V_i \left(\dfrac{\lambda}{2}\mathrm{tr}(\mathbf{G}_i)^2 + \mu \mathrm{tr}(\mathbf{G}_i^2)\right).
\end{align}
Here, $\mathbf{G_i}$ is the Green strain tensor for the face $f_i$: $\mathbf{G_i} = \frac{1}{2}(\mathbf{F_i}^T\mathbf{F_i} - \mathbf{I})$, $V_i$ is the face's volume (thickness$\times$area), and $\lambda$ and $\mu$ are Lam\'e coefficients serving as balancing weights. For  $\lambda$ and $\mu$ we use the same default values as in SNUG~\cite{santesteban2022snug}.

We denote the full physical-regularization term as $\mathcal{L}_{\textit{phys}} = \mathcal{L}_{\textit{bending}} + \mathcal{L}_{\textit{strain}}.$
It helps preserve the physical realism of the tracked mesh but does not provide any information about the underlying body.
Hence, the garments tend to implode and not conform to the body shape.
This issue can be solved using a parametric body mesh fitted to the multi-view sequence.
Following ContourCraft~\cite{grigorev2024contourcraft}, we use cubic energy term to penalize negative normal distance between garment nodes and the body faces closest to them:
\begin{align}
\mathcal{L}_{\textit{body}} = \sum_{i} \mathrm{max}(\epsilon_{\textit{body}} - ((v_i - f_i)\cdot \vec{n}_i), 0)^3,
\end{align}
where $v_i$ are the vertex coordinates, $f_i$ is a point on the body face, $\vec{n}_i$ is this face's normal vector, and $\epsilon_{\textit{body}}$ is a safety margin. 
In our experiments, we set $\epsilon_{\textit{body}}$ to 3mm.

However, for sequences with dynamic body motions, the optimization process often starts far from the target body pose. This large difference in vertex positions causes the optimization to produce unrealistic geometries or diverge completely (see Fig.~\ref{fig:regabl} for illustrations).
We work around this issue by substituting $\mathcal{L}_{\textit{body}}$ with a simple ersatz regularization, $\mathcal{L}_{\textit{VE}}$, in the first half of the optimization process.
This regularization uses virtual edges built between the garment faces opposite each other in the template-frame geometry and penalizes these face pairs for getting too close together (see Sec.~\ref{sec:sm_vedges} for details).
In the second half of the optimization, after the RGB signal pulls the garment geometry to better conform to the body pose, we switch to using the body penetration term $\mathcal{L}_{\textit{body}}$.

The full energy term minimized in the registration process is formulated as
\begin{align}
\mathcal{L}_{\textit{register}} = \lambda_1 \mathcal{L}_{\textit{RGB}} + \lambda_2 \mathcal{L}_{\textit{phys}} + \lambda_3 \mathcal{L}_{\textit{body}},
\end{align}
with $\mathcal{L}_{\textit{body}}$ here substituted by $\mathcal{L}_{\textit{VE}}$ for the first half of the optimization process in each frame.

\subsection{Appearance reconstruction}
\label{sec:appearance}
So far, we have used the Gaussian texture reconstructed from the template frame.
While it provides useful gradients for the registration procedure, its quality is limited by the visual information available in a single time frame. 
Moreover, the lighting conditions are baked into this texture. 
Therefore, we further optimize the garment's appearance using multi-view videos and meshes registered in the previous step.

We disentangle the garment's appearance into two components: a) a base Gaussian texture, introduced in Section~\ref{sec:gtex}. 
b) a texture update predicted by a neural network $f_{\theta}$. This neural network takes as input the albedo occlusion map $A$ and the normal map $N$ of the mesh. 
Following Li et al.~\cite{li2024animatable}, we choose the StyleUNet architecture for $f_{\theta}$.
It predicts offsets to the texture's spherical harmonics  $\Delta \boldsymbol{\phi}$, and translations $\Delta \boldsymbol{\mu}$ in each frame.

\begin{figure*}
    \centering
    \includegraphics[width=1\linewidth]{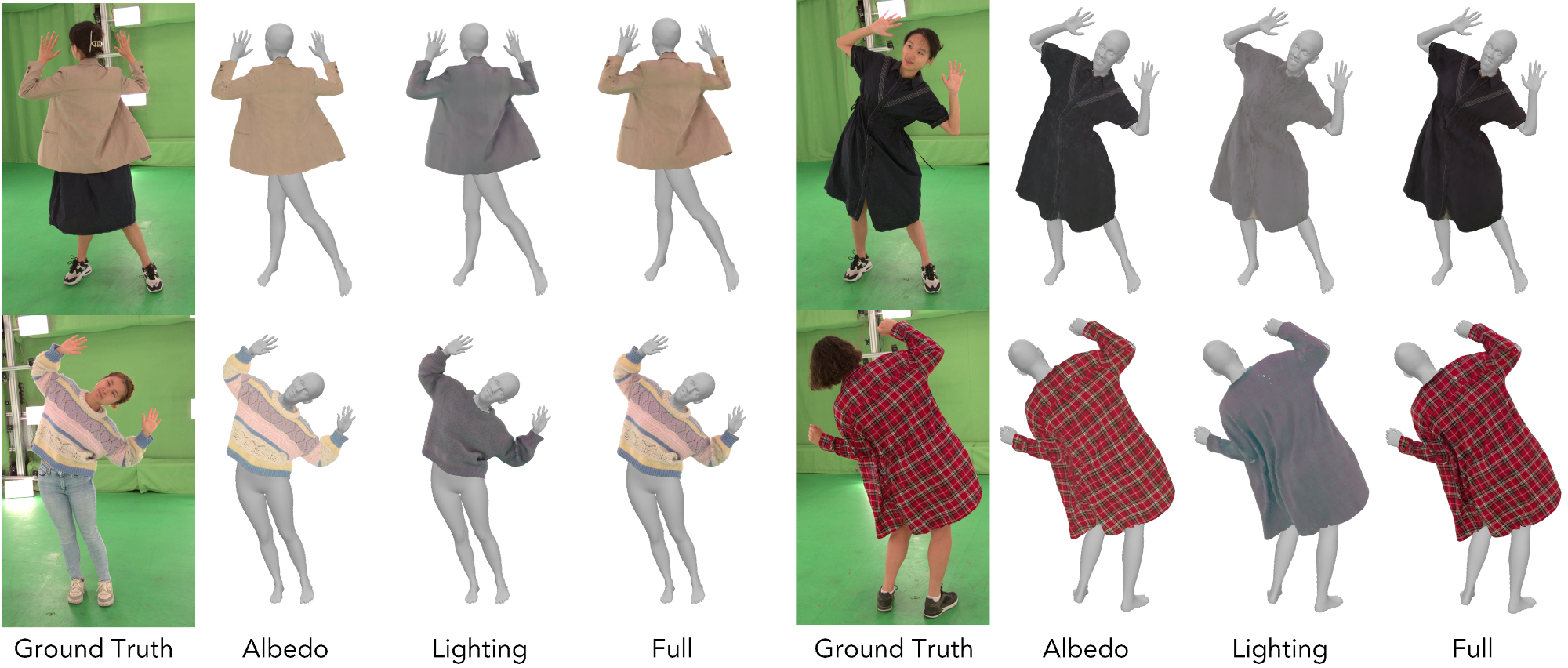}
      \caption{
      We disentangle the albedo color of the Gaussian Garments from the lighting effects predicted by a neural network.
      Here we show four garments rendered over the registered sequence.
      Note how, when rendered with albedo colors, the garments lack any shadows or specular effects.
      The lighting information comes solely from network predictions and matches the ground-truth information.
      The figure shows registered mesh sequences that were not seen by the appearance model during training.}   
      \label{fig:lighting}
\end{figure*}

The predicted offsets to the spherical harmonic coefficients allow the model to separate the albedo colors stored in the base texture from lighting effects (see Fig.~\ref{fig:lighting}).
The translational offsets account for observational noise, preserving high-frequency detail and local geometry of the surface (see Fig.~\ref{fig:appabl} for visual examples).

The final Gaussian texture $\Omega$ for a specific frame is formulated as follows:
\begin{align}
\Omega = \{\boldsymbol{\phi}+\Delta\boldsymbol{\phi}, \alpha, \mathbf{s}, \mathbf{r}, \boldsymbol{\mu}+\Delta\boldsymbol{\mu} \} \in \mathbb{R}^{H\times W \times 59},
\end{align}
where $\Delta\boldsymbol{\phi}$ and $\Delta\boldsymbol{\mu}$ are predicted by $f_{\theta}$:
\begin{align}
\Delta\boldsymbol{\phi}, \Delta\boldsymbol{\mu} = f_{\theta} (A, N).
\end{align}

\subsection{Mesh-based 3DGS rendering}
\begin{figure}
    \centering
    \includegraphics[width=1\linewidth]{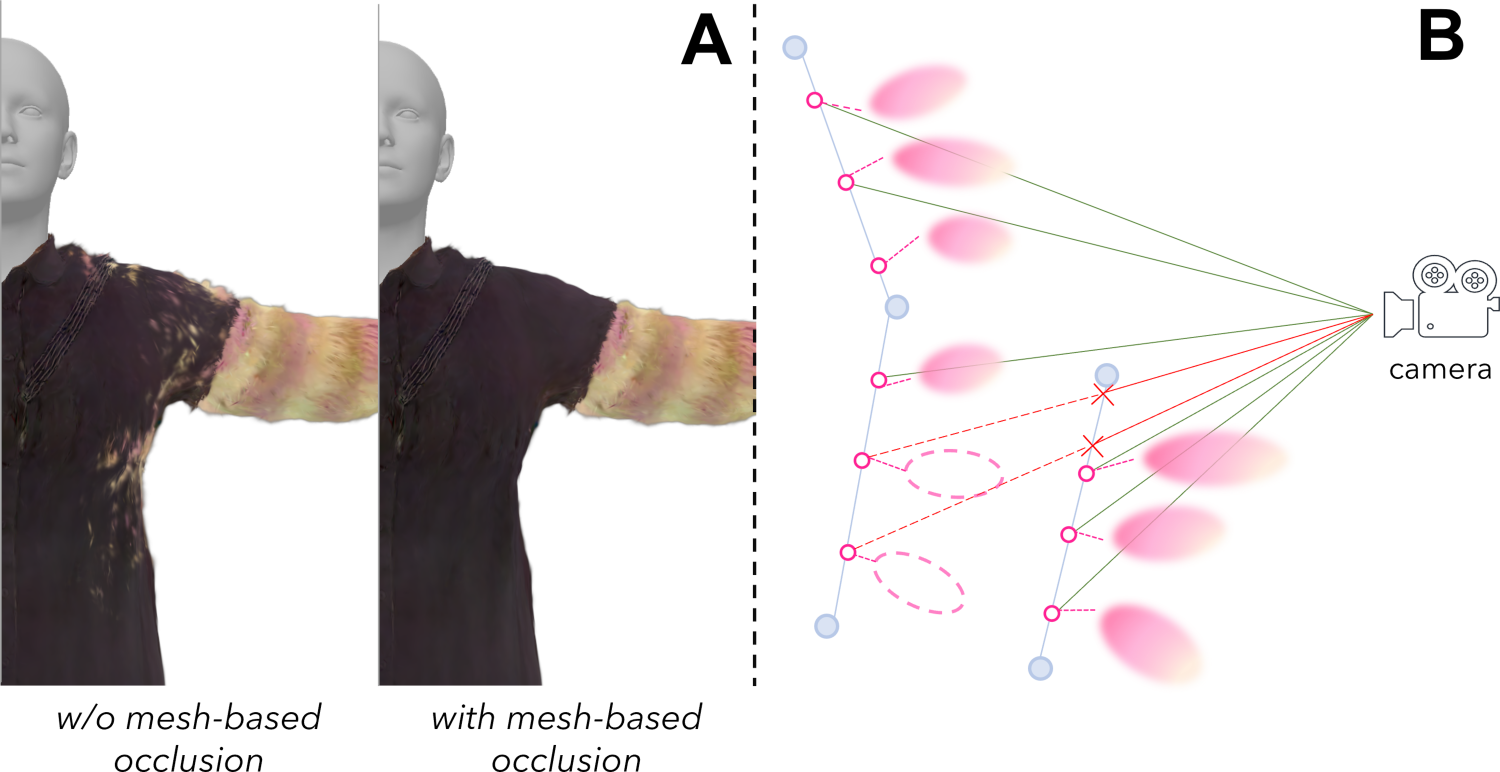}
      \caption{  
      When a ``fuzzy" garment is placed under another one, its Gaussians phase through the outer garment (A, left). 
      We solve this by checking the visibility of the Gaussians' surface points based on the mesh geometries (B).
      Only the Gaussians with visible surface points are rendered (A, right).
      }
      \label{fig:rendering}    
\end{figure}
When modeling surfaces in close proximity using 3D Gaussian splatting, it is crucial to properly handle the visibility of the Gaussians. 
For instance, if a fuzzy surface (e.g., fur) is placed beneath an outer garment layer, the inner layer's Gaussians would incorrectly phase through it (see Fig.~\ref{fig:rendering}A), whereas in reality, the fur would be pressed down by the outer layer.
Properly modeling effects like this would require simulating physical behavior on a per-Gaussian level. 
We address this issue with a simple yet effective workaround that leverages the coupling between the mesh and 3D Gaussian splatting representations.

For each Gaussian, we cast a ray from the camera origin to its corresponding surface point, defined by a mesh face and the point's barycentric coordinates. 
We then check if this point is occluded by another mesh, such as the human body or another garment, and only render the Gaussian if its corresponding surface point is visible (see Fig.~\ref{fig:rendering}B).

\subsection{Behavior fine-tuning}
\label{sec:behavior}
In the final stage of our pipeline, we optimize the garment's behavior.
To simulate the dynamics of Gaussian garments, we employ a learned graph neural network introduced in ContourCraft~\cite{grigorev2024contourcraft}.
This GNN, denoted as $g_{\psi}$, where $\psi$ are the network's parameters, takes as input the nodal positions $\mathbf{x}_t$ and velocities $\mathbf{v}_t$ of the mesh at the current frame $t$, along with each node's material vector $\mathbf{m}$ and each edge's resting geometry $\bar{E}$. 
From these inputs, $g_{\psi}$ predicts the nodal accelerations $\hat{\mathbf{a}}_{t+1}$ for the next frame:
\begin{align}
\hat{\mathbf{a}}_{t+1} = g_{\psi}(\mathbf{x}_t, \mathbf{v}_t, \mathbf{m}, \bar{E})
\end{align}
To fit the observed behavior of the garment, we jointly optimize the model's weights, the material vectors, and the rest edges to minimize the loss function $\mathcal{L}_{\textit{behavior}}$. This loss function combines the mean squared error between the predicted and registered nodal positions with a set of physical terms.

\begin{gather}
\begin{aligned}
\psi^*, \mathbf{m}^*, \bar{E}^* 
 &= \operatorname*{argmin}_{\psi^*, \mathbf{m}^*, \bar{E}^*} \left[\right. \\
 & \sum_t \mathcal{L}_{\textit{behavior}}(g_{\psi}(\mathbf{x}_t, \mathbf{v}_t, \mathbf{m}, \bar{E}), \mathbf{a}_{t+1}) \left.\right]
\end{aligned}
\end{gather}
where $\mathbf{a}_{t+1}$ are the nodal accelerations in frame $t+1$ in a registered sequence. For more details, please refer to Sec.~\ref{sec:sm_behavior}.
\section{Results}
\begin{figure*}
    \centering
    \includegraphics[width=\linewidth]{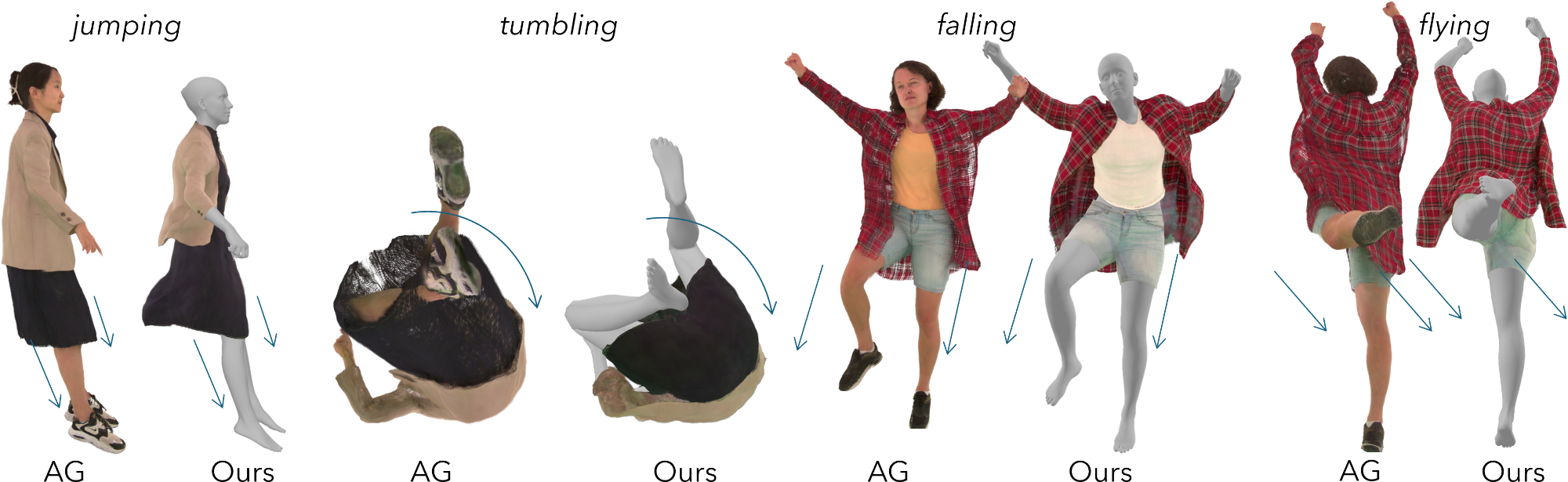}
      \caption{
      Qualitative comparison of our method to Animatable Gaussians~\cite{li2024animatable} (\textit{``AG''}).
      Combining 3DGS with a mesh-based representation, Gaussian Garments are much more robust in simulating challenging poses.
      With learned cloth simulation, we can also more faithfully model dynamic motions.
      }
      \label{fig:vsag}
\end{figure*}
\subsection{Data}
In total, we use 15 garments in our experiments, of which 13 are part of the 4D-Dress dataset~\cite{wang20244d}, and two are newly captured garments with fuzzy fur-like textures. 
The subjects wearing the garments are recorded by 48 cameras regularly placed around them.
Each garment is captured in 6 to 10 video sequences with diverse poses of roughly 150 frames each. 
We use 44 cameras to reconstruct, register, and train the appearance models and validate our results using the remaining 4 cameras.
We train the appearance model and fine-tune the simulation GNN with all multi-view videos available for the given subject except one, holding it out as a validation set.
This way, we evaluate the trained parts of the pipeline (appearance and behavior optimization) on the pose sequences unseen during training.
In our qualitative evaluation and supplementary video, we also use sequences from the AMASS dataset~\cite{mahmood2019amass} demonstrating our ability to generalize to completely new poses and body shapes. 

\begin{figure*}
    \centering
    \includegraphics[width=\linewidth]{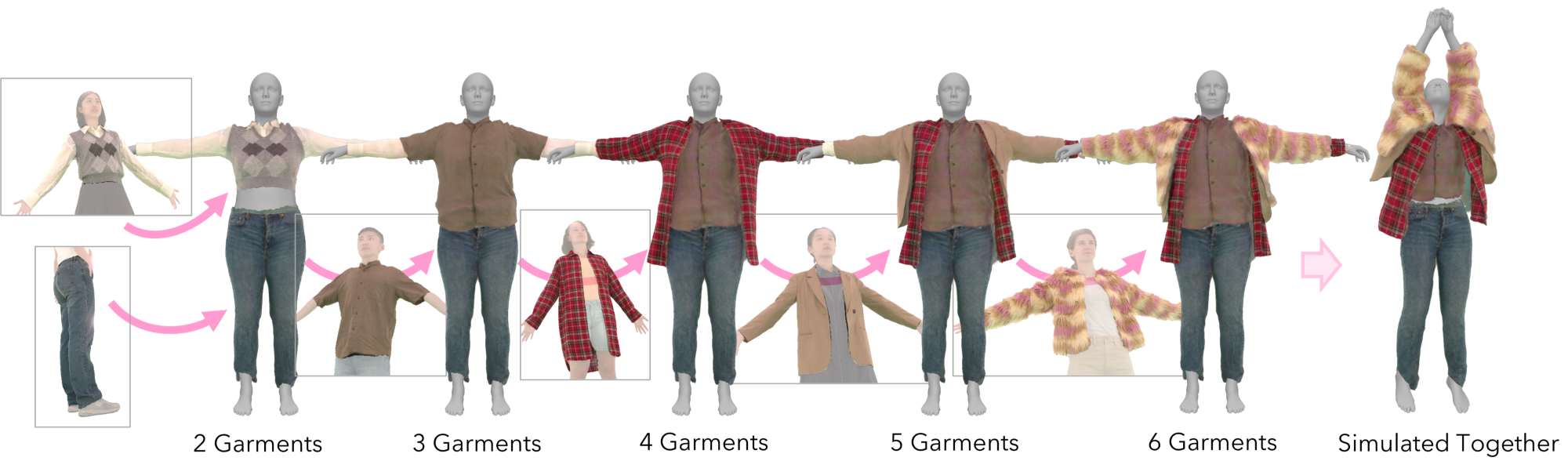}
  \caption{
  We can automatically untangle multiple garments to combine them into a single multi-layer outfit. 
  We then simulate this outfit with the fine-tuned GNN (rightmost).
  }
      \label{fig:multigarment}
\end{figure*}
\subsection{Garment registration}
\begin{table}
\caption{Quantitative ablation study of our registration algorithm.}
\centering
\scalebox{0.8}{
\begin{tabular}{|c|c|c|c|c|}
\hline
                   & F-score, \% $\uparrow$ & CD, cm $\downarrow$ & p2m, cm $\downarrow$  &  $\mathcal{L}_{body}\downarrow$   \\ \hline
\textit{only-RGB}  &   4.4       & 15.5e+2      &  6.19e-1  & 3.47e+2    \\ \hline
\textit{w/o body}  &   88.4      & 1.12   &  5.24e-1  & 4e-3    \\ \hline
\textit{w/ body}   &   89.1      & 6.57   &  5.26e-1  & 4.24e-1    \\ \hline
\textit{Ours-full} &   \textbf{89.6}      & \textbf{1.04}   &  \textbf{5.04e-1}  & \textbf{1.03e-5}    \\ \hline
\end{tabular}
}
\label{tab:regablation}

\end{table}

\label{sec:regable}
We evaluate our algorithm for tracking-based mesh registration.
In this section, we compare it to several ablations. 
In Sec.~\ref{sec:sm_eval_reg} we also compare our registration procedure to the state-of-the-art registration method by Lin et al.~\cite{lin2023leveraging}.
We demonstrate that using multi-view videos our method achieves comparable, albeit slightly lower, accuracy to \cite{lin2023leveraging}, while the latter optimizes template meshes using ground-truth scanned geometries.

For the quantitative analysis, we use three metrics: Chamfer Distance (CD), average point-to-mesh distance (p2m), and F-score \cite{knapitsch2017tanks}, which intuitively describes the percentage of correctly reconstructed points on the mesh surface.
We use a threshold value of 1cm for the F-score.
These metrics measure how close the registration results are to the ground-truth garment meshes, which are reconstructed with a system similar to that used in \cite{collet2015high}. To obtain the individual garment parts, we perform semantic segmentation with the method proposed by Wang et al.~\cite{wang20244d}.
We also compute the body penetration loss $\mathcal{L}_{\textit{body}}$ to measure how well the registered mesh aligns with the underlying body geometry. 

The first ablation \textit{``only-RGB''} optimizes the positions of the mesh vertices using only the RGB signal $\mathcal{L}_{\textit{RGB}}$ without any physics-based regularization. 
In this case, the optimized mesh completely loses its structure, producing disfigured geometry spatially distant from the ground truth (Table~\ref{tab:regablation}).
The second ablation adds two physical terms to the optimization energy: $\mathcal{L}_{\textit{bending}}$ and $\mathcal{L}_{\textit{stretching}}$.
They serve as regularization and help to keep garment geometry physically plausible.
However, the garments optimized without the body geometries tend to implode and do not conform to the observed body.
In Table.~\ref{tab:regablation} we call this ablation \textit{``w/o body.''}
\textit{``w/ body''} uses the body penetration term $\mathcal{L}_{\textit{body}}$ with respect to the parametric body meshes.
This improves the draping in most cases, but in dynamic pose sequences, the optimization may start far away from the next frame body mesh leading to divergence and worse metric values on average.
In our full registration pipeline \textit{``Ours-full,''} we use a substitute loss $\mathcal{L}_{\textit{VE}}$ for the first half of the optimization and then switch back to $\mathcal{L}_{\textit{body}}$ (Sec.~\ref{sec:registration}).
This way, we successfully register the dynamic movement of loose garments like dresses and open jackets (see Fig.~\ref{fig:regabl} for qualitative comparison of the ablations and the supplementary video for further result visuals).

\subsection{Appearance modeling}

\begin{table}[]
\centering
\caption{We quantitatively compare our full appearance model to a set of ablations over the unseen pose sequences and unseen camera views.
Predicting lighting effects and per-frame translation offsets allows us to better match the ground-truth observations.}
\scalebox{0.8}{
\begin{tabular}{|l|l|l|l|}
\hline
                        & LPIPS $\downarrow$ & SSIM $\uparrow$ & PSNR $\uparrow$ \\ \hline
\textit{template-frame} &   1.09e-2    &  0.988    &  36.5    \\ \hline
\textit{only-texture}   &   9.52e-3    &  0.990    &  37.6    \\ \hline
\textit{w/ lighting}    &   8.52e-3    &  0.991    &  38.3    \\ \hline
\textit{Ours-full}      &   \textbf{8.12e-3}    &  \textbf{0.992}    &  \textbf{38.8}    \\ \hline
\end{tabular}
}
\label{tab:appabl}
\end{table}

% Models --- Mean_SSIM --------- Mean_PSNR ----------- Mean_LPIPS
% no_unet   9.90497e-01   3.76131e+01   9.61674e-03
% no_xyz   9.91882e-01   3.82781e+01   8.52511e-03
% fframe   9.88202e-01   3.65451e+01   1.08847e-02
% full   9.92392e-01   3.87641e+01   8.11889e-03
We evaluate the photorealism of our appearance model both quantitatively (Table~\ref{tab:appabl}) and qualitatively (Fig.~\ref{fig:appabl}) by comparing it to several ablations. 
The quantitative evaluation (Table~\ref{tab:appabl}) compares the models in terms of three metrics measuring visual realism: structural similarity (SSIM) \cite{zhou2004image}, learned perceptual similarity (LPIPS) \cite{zhang2018unreasonable}, and peak signal-to-noise ratio (PSNR).
We perform the comparisons using validation videos not seen in training by any of the models and novel camera views.

We first compare our model to a simple \textit{``template-frame''} procedure, which optimizes the Gaussian scene only for the template frame.
This bakes the lighting conditions and any visual artifacts present in the template frame into the garment's appearance.
On the other hand, na\"ively optimizing the Gaussian texture over multiple videos (\textit{``only-texture''}) averages the lighting and high-frequency details, resulting in blurry textures.
The ablation \textit{w/ lighting} optimizes a neural network to predict lighting effects from local information -- ambient occlusion and normal maps.
This helps disentangle the garment's albedo texture from lighting but still averages high-frequency details.
Finally, our full model (\textit{Ours-full}) accounts for the noise in the observations by predicting translational offsets for the Gaussians in each frame, which helps preserve high-frequency information and reduce blur.

We also evaluate our behavior-tuning procedure in Sec.~\ref{sec:sm_eval_beh}.

\subsection{Applications}
\moniker{} create comprehensive representations of real-world garments as distinct 3D assets.
This opens the door for many applications sought by 3D designers.

\begin{figure*}
    \centering
    \includegraphics[width=\linewidth]{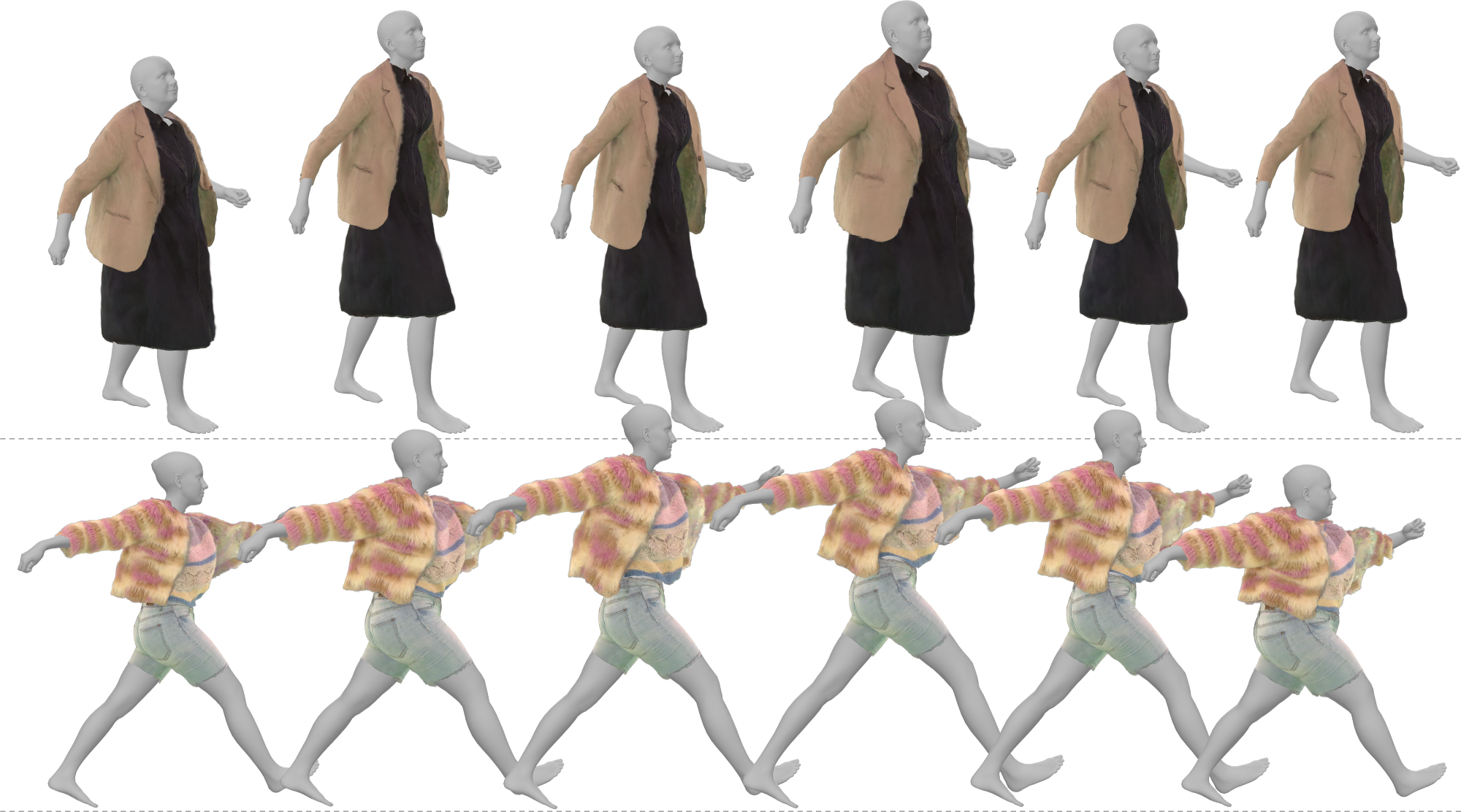}
      \caption{
      We can automatically resize the Gaussian Garments to fit the desired body shape. Here we randomly sample shape parameters for the parametric body model and render the same outfit for different shapes.
      }
      \label{fig:resize}
\end{figure*}
\subsubsection{Simulation}
Gaussian garments can be simulated in dynamic sequences by the fine-tuned ContourCraft~\cite{grigorev2024contourcraft} GNN, which can prevent and resolve cloth self-penetrations, thus automatically modeling re-sized and re-posed multi-layer outfits.
The simulation speed ranges from 10 fps for single garments to ~1 fps for outfits with multiple layers as in Fig.~\ref{fig:multigarment}.

Compared to holistic Gaussian avatars like Animatable Gaussians~\cite{li2024animatable}, using our reconstructed garments with the simulation GNN allows us to robustly model challenging pose sequences like jumps and tumbles. In Fig.~\ref{fig:vsag}  and in the supplementary video, we demonstrate examples where Gaussian Garments excel in modeling the reconstructed outfits compared to~\cite{li2024animatable}. Since our method does not model the non-covered parts of the human body, we do not compare quantitatively to~\cite{li2024animatable}. 
In Sec.~\ref{sec:sm_eval_app} we also provide a quantitative comparison to SCARF~\cite{feng2022capturing}.

\subsubsection{Mix-and-match}
Multiple distinct garment assets, extracted from different multi-view videos, can be combined into novel multi-garment outfits.
We first align each individual garment with the canonical pose and shape of the parametric body model SMPL-X~\cite{pavlakos2019expressive}.
Then, we automatically order the garments by running a simple procedure built around ContourCraft (see Sec.~\ref{sec:sm_ordering}).
The resulting outfit can then be simulated with the fine-tuned ContourCraft model.
In Fig.~\ref{fig:multigarment}, we show how we automatically combine garments into a single simulation-ready outfit.

\subsubsection{Garment resizing}
The reconstructed Gaussian garments and their combinations can be automatically re-sized to match a given body shape, by adjusting the edge lengths in the garments' rest geometry according to the shape blend-weights collected from the SMPL-X body.
We diffuse the body model's blend weights as proposed by Santesteban et al.~\cite{santesteban2021self} to avoid artifacts caused by the resizing.
Fig.~\ref{fig:resize} demonstrates an outfit automatically resized to random body shapes.

\section{Limitations and future work}
While our method can model the overall geometry and photorealistic appearance of garments, the following limitations are to be addressed in future work. 
1) For the appearance model, we assume scenes with uniform lighting. 
Our approach predicts lighting effects based on ambient occlusion and normal maps but does not accommodate dynamic relighting. 
2) While our Gaussian texture can capture high-frequency geometric details like fur to some extent, its effectiveness is limited by the quality of the segmentation masks used during training. 
3) We use Gaussian textures with a fixed resolution of $512^2$ pixels, which may lead to magnification and minification artifacts.
An important direction of future work is adopting standard computer graphics techniques like mipmapping to Gaussian textures.
4) Details such as collars and pockets are represented using the appearance model rather than explicit geometry, as our approach is not aware of the geometry of creases.

Finally, the first three stages of our pipeline can be used with a differentiable physical simulator instead of the learned GNN as long as this simulator allows for optimizing the material parameters of the cloth. We chose ContourCraft~\cite{grigorev2024contourcraft} for its ability to initialize and recover from self-intersecting geometries and its inference speed.

\vspace{-5pt}
\section{Conclusion}

We present \moniker, a comprehensive approach for creating fully controllable 3D clothing assets from multi-view videos based on 3D Gaussian splatting (3DGS). Our approach seamlessly integrates 3DGS with commonly used polygonal meshes to reconstruct the 3D geometry of garments, register it to video observations, optimize garment appearance to achieve photorealistic quality, and fine-tune garment behavior to model dynamic garment motion. We demonstrate results on garment simulation, mixing-and-matching, and resizing as some of the applications of our \moniker. 

\textbf{Acknowledgements.} 
AG was supported in part by the Max Planck ETH Center for Learning Systems. 
We thank Juan Zarate, Wojciech Zielonka, and Peter Kulits for their feedback and help during the project.

\clearpage
\maketitlesupplementary
\appendix
\section{Initial Mesh Reconstruction}
\label{sec:sm_meshinit}

The process of reconstructing the garment mesh from single-frame multi-view imagery involves three key steps.
First, we construct a dense oriented point cloud of the scene by running a multi-view stereo algorithm from COLMAP~\cite{schoenberger2016sfm} using the images of the template frame. 
Next, we filter out background points and reconstruct the surface of the clothed human body using Poisson surface reconstruction~\cite{kazhdan2006poisson}. 
 Finally, we separate the individual garments from the body using semantic segmentation maps and apply a re-meshing algorithm by \cite{levy2013variational} to obtain well-defined triangle meshes of the desired resolution for each garment piece. We use 8000 vertices for each garment, which we observe works well with the pre-trained GNN simulator.
\section{Appearance Details}
\label{sec:sm_appearance}

To model the garment's appearance, we use a Gaussian texture, i.e., a 2D image with multiple channels containing parameters for 3D Gaussians. 
To produce a 3D Gaussian Garment, we sample the Gaussians from the texture in a regular grid (e.g. one Gaussian per texture pixel). Note that each garment face may contain multiple Gaussians depending on the face's texture location. Then, we position the Gaussians in 3D using the sampled parameters. Here we describe this process in detail.

Following the approach of Qian et al.~\cite{qian2024gaussianavatars}, we define a local coordinate system for the Gaussians, which allows us to transform them along with the deforming mesh. The coordinate system for Gaussian $i$ is defined by rotation matrix $\mathbf{R}_j \in \mathrm{SO}(3)$ specific to the face $j$ and the Gaussian surface point $\boldsymbol{\tau}_i \in \mathbb{R}^3$ as the origin. 
The coordinate system is unique for each Gaussian. 
Its basis comprises three unit vectors: the normal vector of the Gaussian's corresponding triangular face, one of the triangle's edges, and the cross-product of these two. 

Inside the coordinate frame, we represent a Gaussian's rotation as a quaternion $\mathbf{r}_i \in \mathbb{H}$, translational offset $\boldsymbol{\mu}_i \in \mathbb{R}^3$, and scale $\mathbf{s}_i \in \mathbb{R}^3$. This allows Gaussians to move within their corresponding mesh face, and to capture high-frequency texture details. Additionally, as the mesh deforms, the Gaussians attached to a face $j$ are affected by the face's scale, $k_j \in \mathbb{R}_{+}$. This scale is computed as $\dfrac{B+H}{2}$, where $B$ and $H$ are the base and height of the triangle.

Then, during rendering the local-frame Gaussians are transformed into world coordinates by the following equations:
\begin{align}
        \mathbf{r}_i' &= \mathbf{R}_j\mathbf{r}_i, \\
        \boldsymbol{\mu}'_i &= k_j\mathbf{R}_j\boldsymbol{\mu}_i + \boldsymbol{\tau}_i, \\
        \mathbf{s}'_i &= k_j\mathbf{s}_i.
\end{align}

\subsection{Appearance Initialization}
\label{sec:sm_appearance_init}
We initialize the appearance using zeros for all Gaussian parameters, create Gaussians on the mesh surface, and optimize them to match the template frame observations.

The primary optimization term here is the RGB error $\mathcal{L}_{\textit{RGB}}$. It combines mean absolute error $\mathcal{L}_1$ and structural similarity error $\mathcal{L}_{\textit{SSIM}}$ between the renders and ground-truth images.
\begin{align}
\mathcal{L}_{\textit{RGB}} = \lambda_{\textit{RGB}}\mathcal{L}_1 + (1-\lambda_{\textit{RGB}})\mathcal{L}_{\textit{SSIM}},
\end{align}
where $\lambda_{\textit{RGB}}$ is a balancing weight.

Additionally, we incorporate two regularization terms introduced by Qian et al.~\cite{qian2024gaussianavatars}. 
The first term, $\mathcal{L}_{\textit{pos}}$, regularizes the Gaussians to stay close to their surface points, defined by their barycentric coordinates.
\begin{align}
\mathcal{L}_{\textit{pos}} = \| \mathrm{max}(\mu - \epsilon_{\textit{pos}}, 0)\|_2,
\end{align}
where $\mu$ are local translations and $\epsilon_{\textit{pos}}$ is a tolerance threshold.

The second term, $\mathcal{L}_{\textit{scale}}$, penalizes the scale $s$ of the Gaussians relative to the underlying mesh triangles.
\begin{align}
\mathcal{L}_{\textit{scale}} = \| \mathrm{max}(s - \epsilon_{\textit{scale}}, 0)\|_2,
\end{align}
where $\epsilon_{\textit{scale}}$ is a tolerance threshold.

We use the resulting set of Gaussians, rigidly attached to the garment surface, to register the mesh to the multi-view videos.

\subsection{Appearance Modelling}

Many recent works model garment appearance on human avatars as a pose-dependent problem. They directly learn a bijective projection from a specific body pose to a certain garment appearance. However, garments' appearances change dynamically. Wrinkle patterns may vary under the same body pose, and different wrinkles may lead to different occlusions and shadows. Minor 3D structures, like fur, also introduce shifts relative to the garment surface. Therefore, we leverage the Style U-Net~\cite{li2024animatable} architecture to predict appearance changes.

Given the deformed mesh in each frame, we first create ambient occlusion and normal maps using the Blender Python library. We used a texture size of 512$\times$512 px$^2$ in our experiments. These two maps provide the occlusion ratios and surface normal information, which helps to learn the shadows and specular effects on the 
 garment surface. 
Then, we concatenate these maps along the color channel. 
We generate ambient occlusion maps separately for both outer and inner garment surfaces to better model their appearance.

The backbone of our model is Style U-Net, a conditional StyleGAN-based~\cite{Karras2018ASG} generator. During training, the model takes as input the ambient occlusion and normal maps together with a view direction map and predicts offsets to the Gaussian texture. Before the forward process, we convert the normal map directions to the camera coordinate. We find it makes training converge faster with better reflective effects. The view direction map is a tensor with the same shape as the normal map. It contains the normalized directions from the camera position to the origin points, converted to the origin points' local coordinates. All invalid texture pixels, which do not correspond to any point on the surfece, are set to zero. The view direction map is first passed through a small CNN with two convolutional layers and then added element-wise to the hidden layer within the Style U-Net. 
The output has 51 channels with the same resolution as the input maps. The first 48 channels are offsets to the spherical harmonics, modeling the lighting effects, including shadows and highlights. The last 3 channels are translational offsets, compensating for registration inaccuracies and shifts of minor 3D structures.

\section{Registration Details}
\label{sec:sm_registration}
Our registration pipeline uses multi-view images as visual guidance and optimizes Gaussian-bounded mesh positions to register the mesh to successive frames. We leverage the RGB and SSIM loss from 3D Gaussian Splatting~\cite{kerbl3Dgaussians} and add physical regularization terms to preserve realistic wrinkles caused by highly dynamic movement. 

However, in cases where large occlusions are present, e.g.,~occlusions by adjacent body parts, the RGB and physical regularizations (bending and stretching) alone do not suffice for convergence, and the mesh tends to implode (see Fig.~\ref{fig:regabl}). To alleviate this problem, we include a body--garment collision term, $\mathcal{L}_{\textit{body}}$, as a further regularizer that provides a displacement constraint when photometric support is lacking.

Still, in highly dynamic motions, the body--garment penetration at the beginning of the optimization procedure can hinder convergence. Therefore, for the first part of the optimization, we substitute $\mathcal{L}_{\textit{body}}$ by a term based on virtual edges, $\mathcal{L}_{\textit{VE}}$, described below.

\subsection{Virtual Edges Regularization}
\label{sec:sm_vedges}
The garment geometry for each frame is initialized at the last frame's converged position.
However, in highly dynamic sequences, the body may move greatly between the frames, resulting in large body--garment penetrations. In these cases, the $\mathcal{L}_{\textit{body}}$ regularizer fails to preserve the garment geometry, which tends to implode. 

Therefore, we construct ``virtual edges'' between opposite faces of the garment mesh to prevent the mesh from collapsing onto itself. We identify such ``opposite'' faces by casting rays along the normal direction of each face and querying for the intersection face. We filter the identified face pairs by only keeping those whose normals are nearly parallel. We compute the following regularization term to prevent the face pairs from getting too close to each other:
\begin{align}
\mathcal{L}_{\textit{VE}} = \sum_i max(L_{e_i} - l_{e_i}, 0)^2,
\end{align}
where $L_{e_i}$ and $l_{e_i}$ are lengths of the edge $e_i$ in the template and the current geometries respectively.

We use $\mathcal{L}_{\textit{VE}}$ in the first half of the optimization and replace it with $\mathcal{L}_{\textit{body}}$ in the second half of the optimization. We observed that this scheduling allows $\mathcal{L}_{\textit{VE}}$ to maintain the mesh structure while $\mathcal{L}_{\textit{RGB}}$ optimizes the mesh node positions. Using $\mathcal{L}_{\textit{body}}$ for the second half of the optimization allows for more accurate physical draping of the garment on the body, providing the best overall results (please see Table~\ref{tab:regablation}).

\subsection{First Frame Matching}
Our dataset consists of multiple multi-view videos of the same scene. For all videos, we start the registration from the same mesh geometry, reconstructed from the template frame. However, the first-frame pose of each video can be very different from the template frame pose in terms of mesh shape and overall position. 

Therefore, we reconstruct a sparse full-body point cloud for the first frame of the new sequences. Then, we find the global rotation and translation that roughly align the template full-body geometry with each video's first frame, by performing an iterative closest point (ICP) algorithm~\cite{EV_ICP} between the point cloud reconstructed from the template frame and the target sequence. 

\section{Behavior Optimization Details}
\label{sec:sm_behavior}
To mimic the real behavior of garments we fine-tune a pre-trained garment simulation GNN from ContourCraft~\cite{grigorev2024contourcraft}. As outlined in the main paper, the GNN autoregressively predicts accelerations $\mathbf{\hat{a}}_{t+1}$ for the mesh nodes in each simulation step given their positions $\mathbf{x}_t$, velocities $\mathbf{v}_t$, material vectors $\mathbf{m}$, and canonical edge lengths $\bar{E}$:
\begin{align}
\hat{\mathbf{a}}_{t+1} = g_{\psi}(\mathbf{x}_t, \mathbf{v}_t, \mathbf{m}, \bar{E})
\end{align}

The geometry for each step is computed by integrating the predicted accelerations into the simulation:
\begin{align}
\hat{\mathbf{x}}_{t+1} = \mathbf{x}_{t} + \mathbf{v}_{t} +  \hat{\mathbf{a}}_{t+1}
\end{align}
For simplicity, we assume a time difference equal to 1 between successive frames.

Our goal here is to optimize $\psi$, $\mathbf{m}$, and $\bar{E}$ so that our simulations better match the behavior of the registered sequences. $\psi$ are the parameters of the GNN and $\mathbf{m}$ are material vectors. These are 4-value vectors attached to each node of the garment mesh and fed into the GNN. $\bar{E}$ are canonical lengths of each edge in the mesh represented by scalar values. All these elements are parts of the original ContourCraft model that we optimize for our needs.

We tune these parameters using all the registered sequences in our training set. During fine-tuning we autoregressively simulate each training sequence with $g_\psi$. In each frame, we use the simulated geometry $\hat{\mathbf{x}}_{t+1}$ to compute a loss value which comprises two terms:
\begin{align}
\mathcal{L}_{\textit{behavior}}(\hat{\mathbf{x}}_{t+1}, \mathbf{x}_{t+1}) &= \mathcal{L}_{\textit{ccraft}}(\hat{\mathbf{x}}_{t+1}, \mathbf{x}_{t+1}) \\&+ \lambda\mathcal{L}_{2}(\hat{\mathbf{x}}_{t+1}, \mathbf{x}_{t+1}),
\end{align}
where $\mathbf{x}_{t+1}$ is the registered geometry for the frame $t+1$, and $\lambda$ is a balancing weight. $\mathcal{L}_{\textit{ccraft}}$ here is the original loss function from ContourCraft, while $\mathcal{L}_{2}$ is a simple mean squared error. 

Finetuning the GNN  using only registered sequences and $\mathcal{L}_{\textit{behavior}}$ enables it to mimic the behavior of the garments. The problem, however, is that our registered sequences only contain individual garments without multi-layer outfits. Because of this, the ContourCraft GNN, which originally could handle multi-layer outfits, starts forgetting how to properly handle multi-layer structures during fine-tuning. To alleviate this issue, we construct a set of multi-layer outfits from our reconstructed garments and use them in every other training iteration instead of the registered individual garments. Since we don't have target geometries for these outfits, we only supervise these steps with $\mathcal{L}_{\textit{ccraft}}$. This enables the model to both match the real garment behavior and properly handle inter-layer collisions.

\section{Automatic Garment Ordering Procedure}
\label{sec:sm_ordering}
We use the ContourCraft~\cite{grigorev2024contourcraft} GNN to devise a simple procedure to automatically untangle and order individual garments.

We start with all garments aligned with the canonical SMPL-X pose and shape. We order the garments by their position in the desired outfit---from the innermost to the outermost. Then we untangle each subsequent garment from the ones that should be beneath it, see Alg.~\ref{algo:untall}.

To untangle a single garment we run two consecutive simulation stages. 
In the first one, we treat all the inner garments as solid bodies. This way, ContourCraft treats them as body geometry and pushes the outer garment outside them.
Then, in the second stage, we simulate all garments, treating them as cloth.
We repeat this procedure twice. See Alg.~\ref{algo:untone}

The whole process takes around 1 minute for each garment on an NVIDIA GeForce 4090 GPU.

\begin{algorithm}
\SetKwFunction{untangle}{Untangle}
\DontPrintSemicolon
\caption{$UntangleAll$; we untangle a sequence of garments one by one from the innermost to the outermost.}
\label{algo:untall}
    % \SetAlgoLined
    \KwInput{Garment geometries $G_1 \ldotp\ldotp G_N$ ordered from the innermost to the outermost}

    \For{$i \in [2\ldotp\ldotp N]$}{
        $G_{outer} \leftarrow G_i$
        
        $G_{inner} \leftarrow [G_{1}\ldotp\ldotp G_{i-1}]$
        
        \untangle($G_{outer}$, $G_{inner}$)
    }

    \Return $G_1 \ldotp\ldotp G_N$
\end{algorithm}

\begin{algorithm}
\SetKwFunction{simulate}{Simulate}
\DontPrintSemicolon
\caption{$Untangle$; to untangle a single garment, we first simulate it over inner ones treating the latter as solid bodies. Then we re-simulate all the garments together as cloth. We set $N_{epochs}$ to 2}
\label{algo:untone}
    % \SetAlgoLined
    \KwInput{Outer garment $G_{outer}$; set of inner garments $G_{inner}$}

    \For{$i \in [1\ldotp\ldotp N_{epochs}]$}{
        $G_{cloth} \leftarrow G_{outer}$
        
        $G_{solid} \leftarrow G_{inner}$
        
        \simulate($G_{cloth}$, $G_{solid}$)

        $G_{cloth} \leftarrow  G_{inner} + \{G_{outer}\}$
        
        \simulate($G_{cloth}$, $\emptyset$)
    }

    \Return $G_{outer}$, $G_{inner}$

\end{algorithm}
\section{Additional Evaluation}
\label{sec:sm_eval}
\subsection{Registration}
\label{sec:sm_eval_reg}
\begin{figure*}
    \centering
    \includegraphics[width=1.0\linewidth]{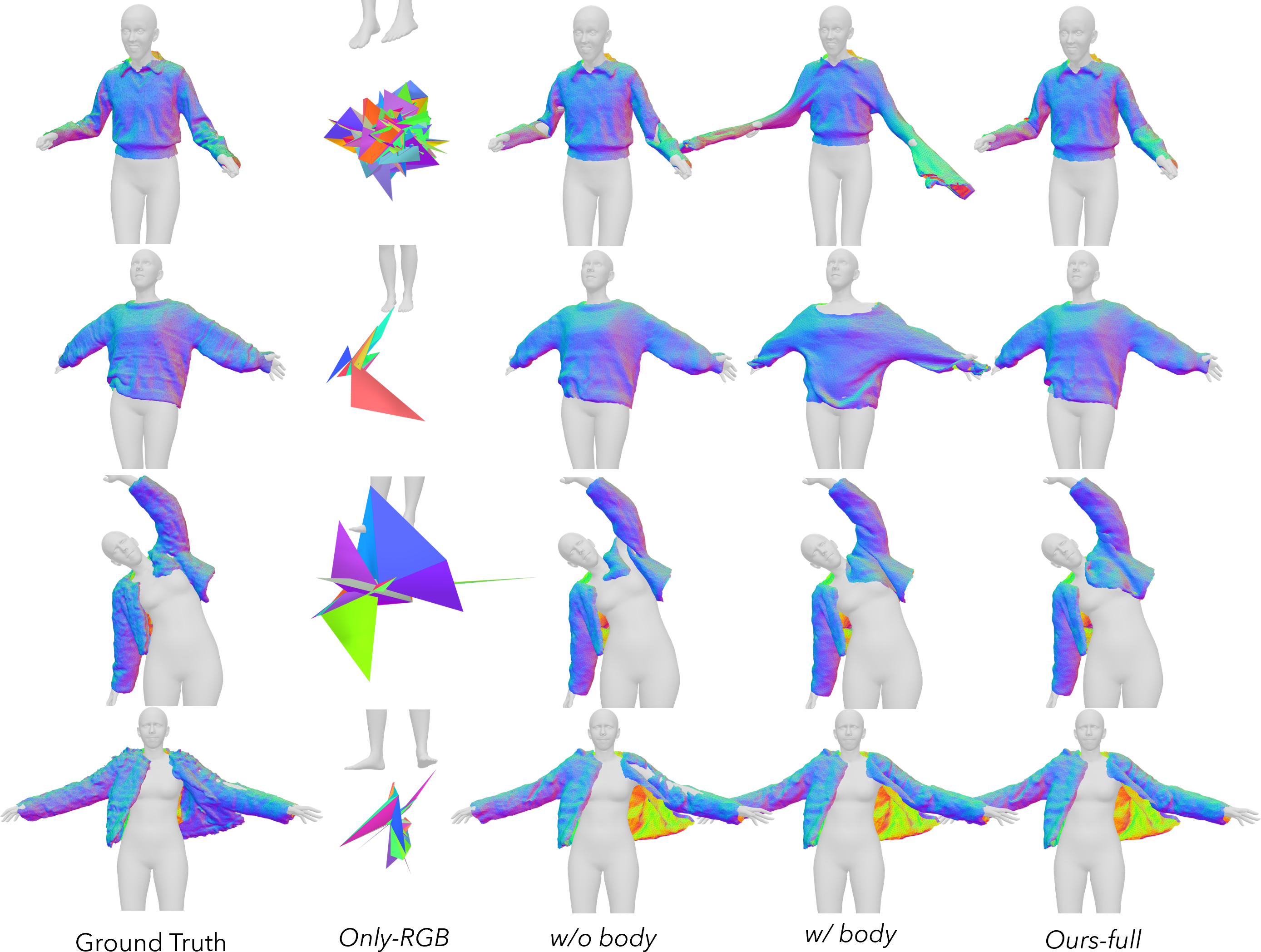}
  \caption{   
  Qualitative comparison of our full registration algorithm and the ablations.
  When only optimizing the RGB loss (\textit{only-RGB}), the optimization diverges completely.
  With physical losses (\textit{w/o body}) the garment preserves its structure bur does not always conform to the body.
  When using the body penetration term (\textit{w/ body}), the optimization if prone to artifacts caused by the incorrect initialization.
  With our full pipeline (\textit{Ours-full}) we first pull the garment geometry closer to the body pose and then enable the body penetration term.
  }
      \label{fig:regabl}
\end{figure*}
We present qualitative examples of our registration ablations in Fig.~\ref{fig:regabl}. When using only the RGB loss (``\textit{Only-RGB}'') the garment geometry diverges within a few steps. This is because optimizing 3D geometry using solely the RGB loss is an ill-posed problem, especially with monochromatic objects, like many garments in our dataset. Moreover, the renders that use the base Gaussian texture may not exactly match GT frames, resulting in noisy signals that accumulate over several frames and lead to diverging results. Introducing the physics losses without an underlying body geometry (``\textit{w/o body}'') has a regularizing effect preventing physically implausible results. However, large body--garment penetrations occur. 
Na\"ively penalizing body--garment collisions (``\textit{w/ body}'') does not allow for robust optimization because the collision computation cannot handle fast movements due to bad initialization.
For instance, if a hand goes through the sleeve between time frames, the body collision term will push the sleeve outside the body instead of pulling it back on.
Therefore, we demonstrate that our full model (``\textit{Ours-full}'') works best for all pose sequences.

\begin{table}
\caption{Comparison between our registration stage and Lin et al.~\cite{lin2023leveraging}. Our method only uses multi-view RGB images as supervision, whereas \cite{lin2023leveraging} directly optimizes a template mesh to fit GT scans.}
\centering
\scalebox{0.9}{
\begin{tabular}{|c|c|c|c|c|}
\hline
                   & F-score, \% $\uparrow$ & CD, cm $\downarrow$ & p2m, cm $\downarrow$  &  $\mathcal{L}_{body} \downarrow$   \\ \hline
\cite{lin2023leveraging}  &   \textbf{98.7}       & \textbf{0.399}      &  \textbf{0.134}  & 4.60e-5    \\ \hline
\textit{Ours}  &   90.4      & 1.001   &  0.486  & \textbf{1.24e-5}    \\ \hline
\end{tabular}
}
\label{tab:regComp}

\end{table}

\begin{figure*}
    \includegraphics[width=\linewidth]{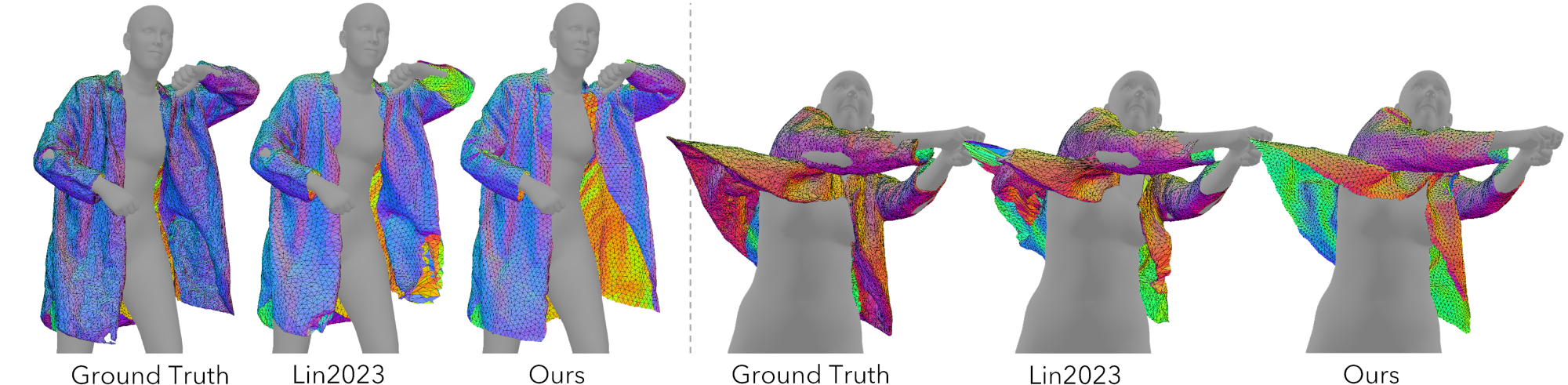}
      \caption{
      Qualitative comparison of our method to Lin et al.~\cite{lin2023leveraging} (\textit{``Lin2023''}).
      Since Lin et al. register garments to ground-truth scans, it may overfit the artifacts present in these scans.
      In contrast, our registration procedure only uses multiview RGB videos and produces physically realistic geometries.
      }
      \label{fig:regComp}
\end{figure*}

We also compare to a state-of-the-art method for garment registration by Lin et al.~\cite{lin2023leveraging} (\textit{``Lin2023''}), using 13 garments from the 4D-Dress dataset.
While our method relies only on multi-view observations from RGB cameras,  \cite{lin2023leveraging} fit the garment template to the same GT scans as used for evaluation.
Given this, our registration procedure performs only slightly worse than \cite{lin2023leveraging} (see Table~\ref{tab:regComp}). Our method performs only 0.6 cm worse in terms of Chamfer Distance (CD) and 0.2 cm worse in point-to-mesh distance. Meanwhile, the scan data in 4D-Dress dataset usually contains outlier faces, e.g.,~closed dress bottoms or duplicate layers on two sides of an open jacket. Given the large data volume, removing all erroneous structures from the ground-truth data is difficult. As a result, \cite{lin2023leveraging} overfits these artifacts leading to faulty geometries (Fig.~\ref{fig:regComp}).

\subsection{Appearance}
\label{sec:sm_eval_app}
\begin{figure*}
    \centering
    \includegraphics[width=\linewidth]{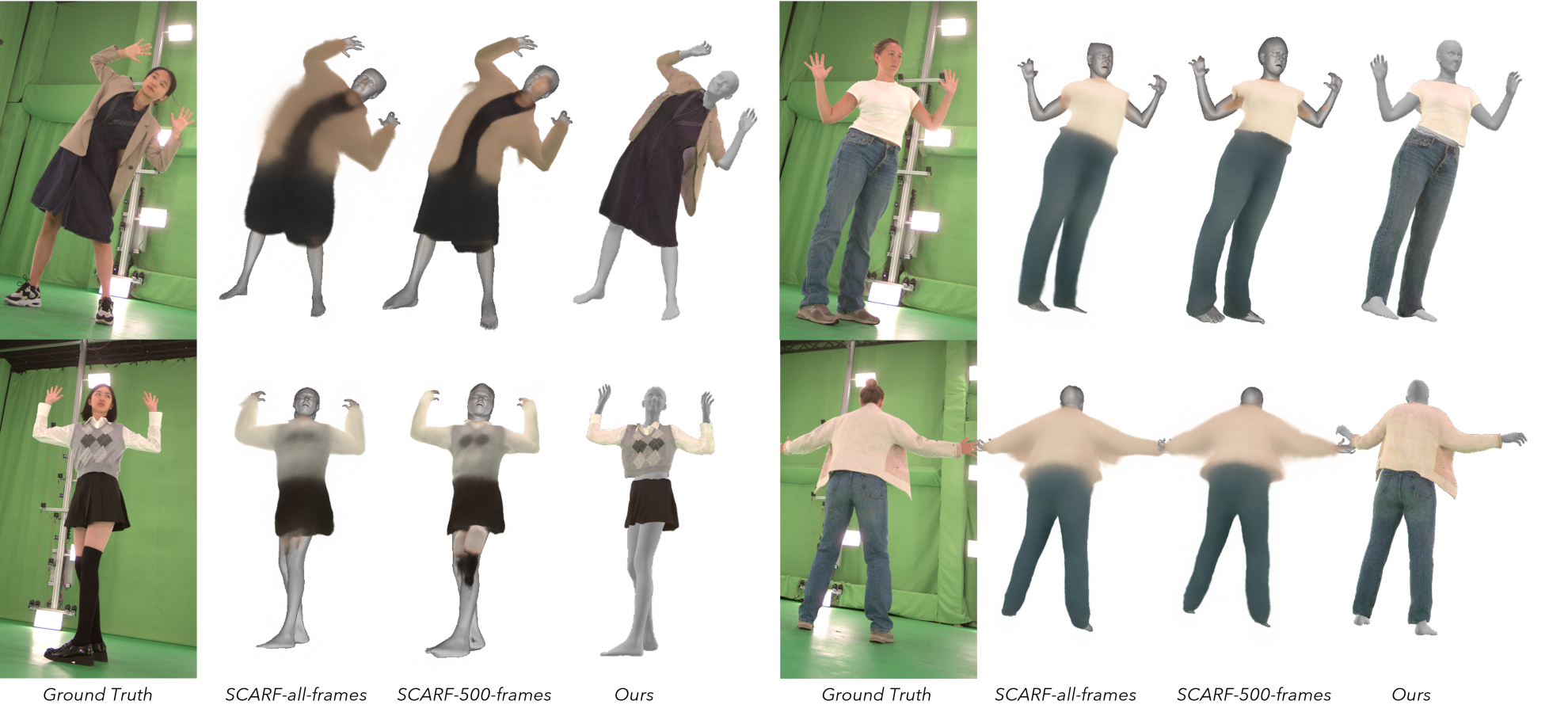}
      \caption{
      Visual comparison of our method to SCARF~\cite{feng2022capturing}. 
      The appearance model and a fine-tuned garment simulation GNN enable Gaussian Garments to produce visually appealing results and better model garment dynamics.
      The body meshes shown above are included for visualization only and were not used in the quantitative evaluation in Table~\ref{tab:vsscarf}.
      SCARF also optimizes offsets to the body geometry resulting in slightly different body models compared to the original SMPL-X used by Gaussian Garments.
      }
      \label{fig:vsscarf}
\end{figure*}

\begin{figure*}[h!]
    \centering
    \includegraphics[width=\linewidth]{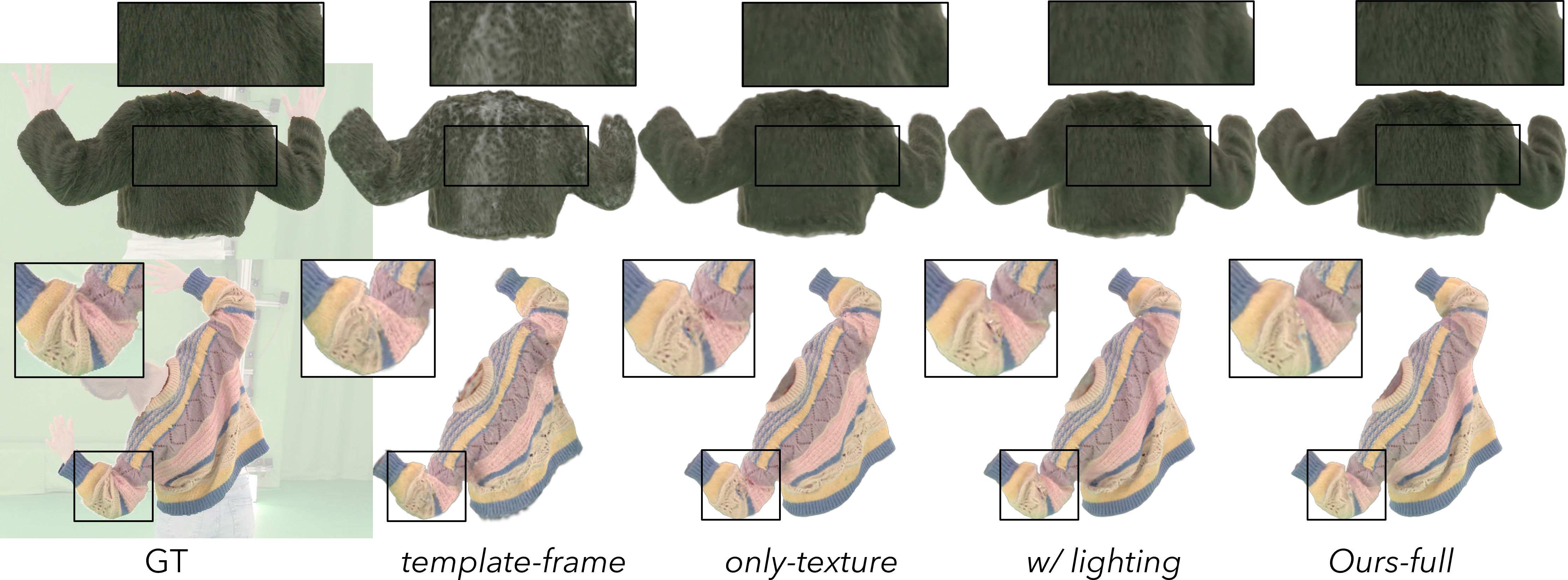}
  \caption{   
  Qualitative comparison of our full appearance model to a sequence of ablations. 
  Note how our full model preserves more high-frequency details and does not contain lighting artifacts.
  }
      \label{fig:appabl}
\end{figure*}

In Fig.~\ref{fig:appabl} we show a visual comparison of our final appearance model to the set of ablations described in the main paper.
On top of this, we also compare our method to SCARF~\cite{feng2024gaussian}.
SCARF is a NeRF-based method that reconstructs an articulated garment radiance field from a \textit{monocular} video.
While SCARF is not a direct baseline to our method due to it using only monocular data, it is the closest method to ours which has publically available code. For this comparison, we use four outfits created from individual Gaussian garments that match the outfit of each subject (see Fig.~\ref{fig:vsscarf}). To optimize the SCARF model we concatenate training frames from different videos and different cameras and treat them as monocular videos. We find that if optimized over frames from all videos and all cameras, SCARF produces extremely blurry results due to data stochasticity. We call the models optimized over all frames \textit{``SCARF-all-frames''}. We also optimize SCARF models over only 500 frames from our videos, making sure they cover the whole body surface in diverse poses. We call such models \textit{``SCARF-500-frames''}. 
The models optimized over 500 frames produce much crisper results but still do not match the ground truth as well as those of Gaussian Garments. 
Please see Fig.~\ref{fig:vsscarf} for visual comparison and Table~\ref{tab:vsscarf} for quantitative evaluation.

\begin{table}[]
\centering
\caption{Quantitative comparison of our method against SCARF. Gaussian Garments' appearance model and fine-tuned garment simulation GNN allow it to produce high-quality visuals that align with ground-truth observations.}
\scalebox{0.9}{
\begin{tabular}{|r|c|c|c|}
\hline
                        & LPIPS $\downarrow$ & SSIM $\uparrow$ & PSNR $\uparrow$ \\ \hline
\textit{SCARF-all-frames}   &   7.65e-2    &  0.884    &  37.3    \\ \hline
\textit{SCARF-500-frames}    &   6.78e-2    &  0.928    &  39.2    \\ \hline
\textit{Ours}      &   \textbf{4.80e-2}    &  \textbf{0.951}    &  \textbf{41.5}    \\ \hline
\end{tabular}
}
\label{tab:vsscarf}

\end{table}


\subsection{Behavior}

\label{sec:sm_eval_beh}
\begin{table}
\caption{Quantitative evaluation of our behavior-tuning procedure. We compare sequences simulated by the GNN to the registered sequences using the L2 loss term. By fine-tuning the garment simulation GNN, our method can match the behavior of the registered and ground-truth meshes more closely.}
\label{tab:material}
\centering
\scalebox{1}{
\begin{tabular}{|r|c|}
\hline
\multicolumn{1}{|l|}{}  & L2  $\downarrow$                   \\ \hline
\textit{default}        & 5.43e-2               \\ \hline
\textit{tuned-leave-one-out}    & 4.38e-2  \\ \hline
\textit{tuned-together} & \textbf{4.34e-2}               \\ \hline
\end{tabular}
}
\end{table}

\begin{figure*}
    \centering
    \includegraphics[width=\linewidth]{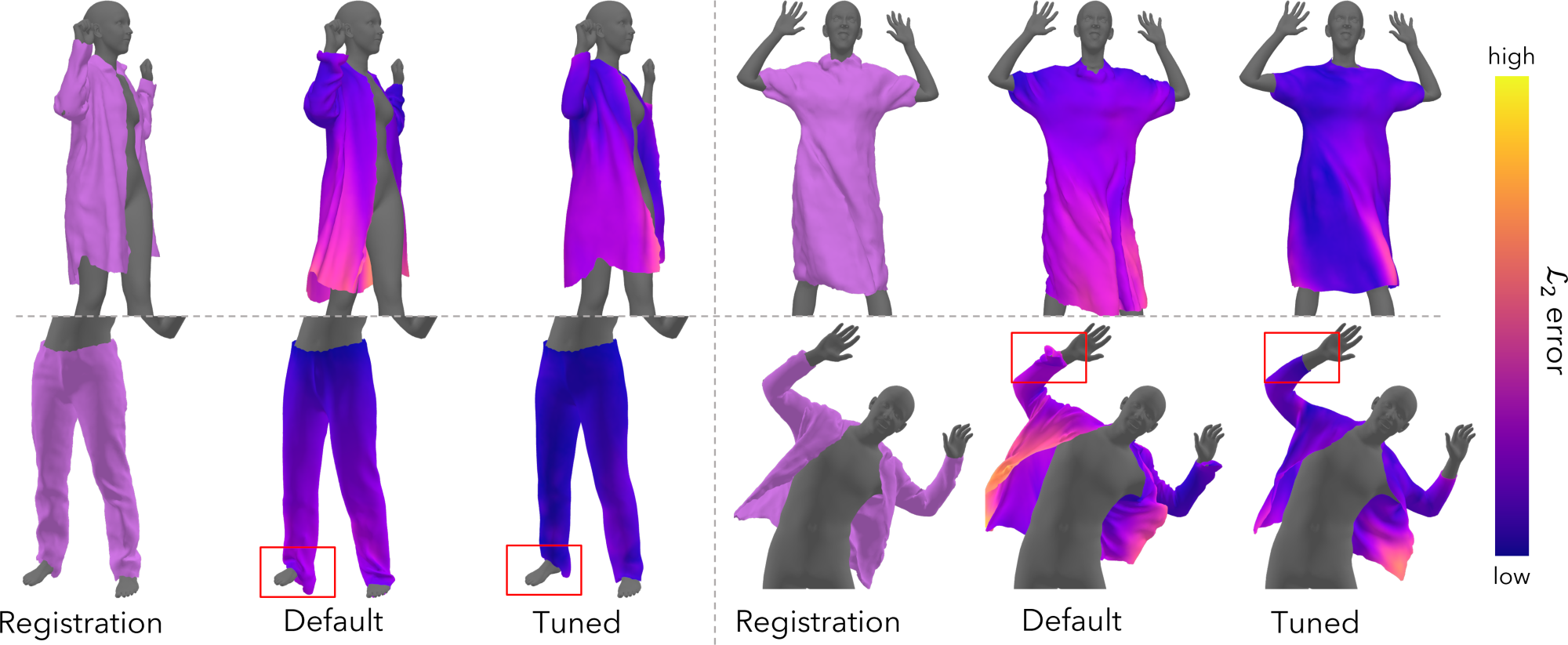}
      \caption{
        Visual comparisons of the simulations produced by a default pre-trained ContourCraft~\cite{grigorev2024contourcraft} model and our fine-tuned version.
        A brighter color denotes a higher L2 error between the simulation and the registered mesh.
        The fine-tuned model achieves behavior that better matches the registered sequences.
        By optimizing the rest geometries of the garments we also better match the original size of the garments (bottom left) and avoid simulation artifacts (bottom right).
      }
      \label{fig:material}
\end{figure*}
Here we evaluate the efficiency of our behavior reconstruction procedure.
To do that, we fine-tune the garment modeling GNN from~\cite{grigorev2024contourcraft} and optimize per-vertex material vectors together with rest geometries over the training sequences.
We then simulate the garments for the held-out sequences and compare them to garment registrations obtained by our approach using the mean L2 distance between the simulated and registered vertex positions.  

In Table~\ref{tab:material} we compare the \textit{``default''} untuned GNN to two tuned variants. In both variants, we optimize the GNN parameters $\psi$ together with the material vectors $\mathbf{m}$ and rest edge lengths $\bar{E}$ for the garments. We call the latter two ``garment parameters''. In \textit{``tuned-together''} we optimize the network parameters and the garment parameters for all 15 garments together and then run an evaluation on the validation sequences. Then, we use the \textit{``tuned-leave-one-out''} variant to demonstrate how a fine-tuned GNN can generalize to garments that are not in the original fine-tuning set. Here we finetune a separate model for each garment in two stages. In the first stage, we optimize the model parameters and garment parameters for all garments except one. In the second stage, we freeze the model parameters and only optimize the garment parameters for the remaining unseen garments. This results in 15 models---one for each garment. We evaluate each model using the validation sequence for the remaining left-out garment. As seen from Table~\ref{tab:material}, models from \textit{``tuned-leave-one-out''} perform only slightly worse than the one from \textit{``tuned-together''}. Hence, we can expect reasonable results for novel garments without fine-tuning the GNN parameters again.

\subsection{Applications}

We demonstrate results in the following applications: simulating the garments in novel and dynamic poses, mixing and matching, and dynamic resizing. 

In Fig.~\ref{fig:vsag} we show a qualitative comparison of our method to AnimatableGaussians~\cite{li2024animatable} for novel and dynamic pose sequences. Our method manages to realistically capture garment motions in dynamic scenes. 

We further demonstrate garment mix-and-match in Fig.~\ref{fig:resize}, where we combine garments from two (top) and three (bottom) different subjects, and automatically resize them to fit diverse body shapes.

Additional results and animated sequences are provided in the supplementary video.

\subsection{Reconstruction time}
We reconstruct each garment separately. In our experiments, we use 1050 multi-view frames with 44 camera views to reconstruct each garment. Our registration and appearance optimization procedures take roughly 36 hours on an NVIDIA GeForce RTX 2080 Ti. 
Specifically, it takes 3.5 hours to register each sequence (on average 150 frames) and 24.5 hours for all sequences. Afterward, it takes 1.5 hours to create ambient occlusion and normal maps in Blender, and 10 hours to train appearance models with 5 epochs on 44 camera view data (46200 images in total). 
It takes an additional ~20 hours for the behavior finetuning stage.

{
    \small
    \bibliographystyle{ieeenat_fullname}
    \bibliography{bibliography}
}
\end{document}